%% file: main.tex
\documentclass{article}

\usepackage{microtype}
\usepackage{graphicx}
\usepackage{subfigure}
\usepackage{booktabs} 

\usepackage{hyperref}
\usepackage{diagbox}
\usepackage{tikz}
\usetikzlibrary{shapes,arrows}

\usepackage[accepted]{icml2024}

\usepackage{amsmath}
\usepackage{amssymb}
\usepackage{mathtools}
\usepackage{amsthm}

\usepackage[capitalize,noabbrev]{cleveref}

\theoremstyle{plain}

\theoremstyle{definition}

\theoremstyle{remark}

\DeclareMathOperator*{\argmax}{arg\,max}

\usepackage[textsize=tiny]{todonotes}

\icmltitlerunning{Hierarchical Transformers are Efficient Meta-Reinforcement Learners}

\begin{document}

\twocolumn[
\icmltitle{Hierarchical Transformers are Efficient Meta-Reinforcement Learners}



\icmlsetsymbol{equal}{*}

\begin{icmlauthorlist}
\icmlauthor{Gresa Shala}{yyy}
\icmlauthor{André Biedenkapp}{yyy}
\icmlauthor{Josif Grabocka}{comp}

\end{icmlauthorlist}

\icmlaffiliation{yyy}{University of Freiburg}
\icmlaffiliation{comp}{University of Technology Nuremberg}

\icmlcorrespondingauthor{Gresa Shala}{shalag@cs.uni-freiburg.de}

\icmlkeywords{Machine Learning, ICML, Reinforcement Learning, Meta-Learning}

\vskip 0.3in
]



\printAffiliationsAndNotice{}  

\begin{abstract}
We introduce Hierarchical Transformers for Meta-Reinforcement Learning (HTrMRL), a powerful online meta-reinforcement learning approach.
HTrMRL aims to address the challenge of enabling reinforcement learning agents to perform effectively in previously unseen tasks.
We demonstrate how past episodes serve as a rich source of information, which our model effectively distills and applies to new contexts.
Our learned algorithm is capable of outperforming the previous state-of-the-art and provides more efficient meta-training while significantly improving generalization capabilities.
Experimental results, obtained across various simulated tasks of the Meta-World Benchmark, indicate a significant improvement in learning efficiency and adaptability compared to the state-of-the-art on a variety of tasks.
Our approach not only enhances the agent's ability to generalize from limited data but also paves the way for more robust and versatile AI systems.
\end{abstract}

\section{Introduction}
\input{1_intro}
\section{Related Work}
\input{2_related_work}
\section{Preliminaries}
\input{3_preliminaries}
\input{4_method}
\section{Experiments}
\input{5_research_questions_results}
\section{Conclusion}
\input{6_conclusion}

\clearpage
\section*{Acknowledgements}
The authors acknowledge funding by The Carl Zeiss Foundation through the research network "Responsive and Scalable Learning for Robots Assisting Humans" (ReScaLe) of the University of Freiburg.
\section*{Broader Impact} This paper presents work whose goal is to advance the field of Machine Learning. There are many potential societal consequences of our work, none of which we feel must be specifically highlighted here.
\bibliography{icml_2024}
\bibliographystyle{icml2024}

\newpage
\appendix
\onecolumn
\section{Appendix}
\input{7_appendix}


\end{document}

%% file: 1_intro.tex
\label{sec:intro}

Reinforcement learning \citep[RL;][]{rl-book2018} has made rapid progress in recent years and has shown success in complex real-world benchmarks \citep{bellemare-nature20a,degrave-nature22a,kaufmann-nature23a} that go beyond typical (video) game playing benchmarks \citep{mnih-nature13,schrittwieser-nature20a}.
Besides such high-profile success stories, RL has not yet found widespread adoption in many real-world applications.
In part, this is due to the high computational demands \citep{cobbe-icml20a,shala-metalearn22a} and the brittleness of RL algorithms \citep{henderson-aaai18a,engstrom-iclr20a,andrychowicz-iclr21a}, which is further amplified by the difficulty of optimizing the hyperparameters of RL algorithms \citep{parker-holder-jair22a,eimer-icml23a,mohan-automl23a}.
However, more importantly, RL training pipelines are generally not built with generalization in mind \citep{kirk-jair23a}.

The lack of generalization capabilities is a crucial factor that limits the applicability of RL in domains where there are no perfect simulators.
The classical RL training pipeline involves training, validating, and testing in the exact same task.
However, in reality, an RL agent's sensors might slightly drift over time or tasks might exhibit drastically different features during different stages of an episode (such as, e.g., times of the day).
If the training task was not set up with such changes in mind, an RL policy will not have the opportunity to learn generalizable behavior.
Consequently, small variations in the task that might not have been observed during training can lead to severe failure cases.
For example, a policy that was trained to steer a robot in a highly controlled lab environment might fail the instant it is supposed to leave the familiar lab environment.

Meta-learning \citep{vanshoren-automlbook19,hospedales-tpami22a} is a subfield of machine learning that aims to take advantage of previous experience as efficiently as possible so that learned behavior can be transferred to similar learning problems.
For example, one avenue of meta-learning that has shown success in RL is to include meta-features directly in the training procedure \citep{zhang-iclr21a,eimer-ijcai21,benjamins-tmlr23a,beukman-neurips23a} to learn policies capable of zero-shot generalization, based on observed meta-features.
Still, it might not always be possible to provide meaningful meta-features to learn generalizable policies.
Instead, a different avenue of meta-learning aims at directly learning how to efficiently learn on new, potentially unseen tasks.
Meta-RL, combines both RL and meta-learning to efficiently learn to reinforcement learn \citep{beck2023survey}.
Here, we present a novel transformer architecture that improves the adaptability of learned RL agents across diverse sets of tasks.
While prior approaches using transformer architectures only aimed at exploiting intra-episode experiences, our approach efficiently leverages intra-episode experiences as well as inter-episode experiences.
Our learned algorithm is capable of outperforming the previous state-of-the-art and providing more efficient meta-training while significantly improving generalization capabilities.
To foster reproducibility, we provide our code at \url{https://github.com/releaunifreiburg/HT4MRL.git}

%% file: 2_related_work.tex
\label{sec:related_work}
\begin{figure*}[ht!]
    \centering
    \includegraphics[width=0.95\textwidth]{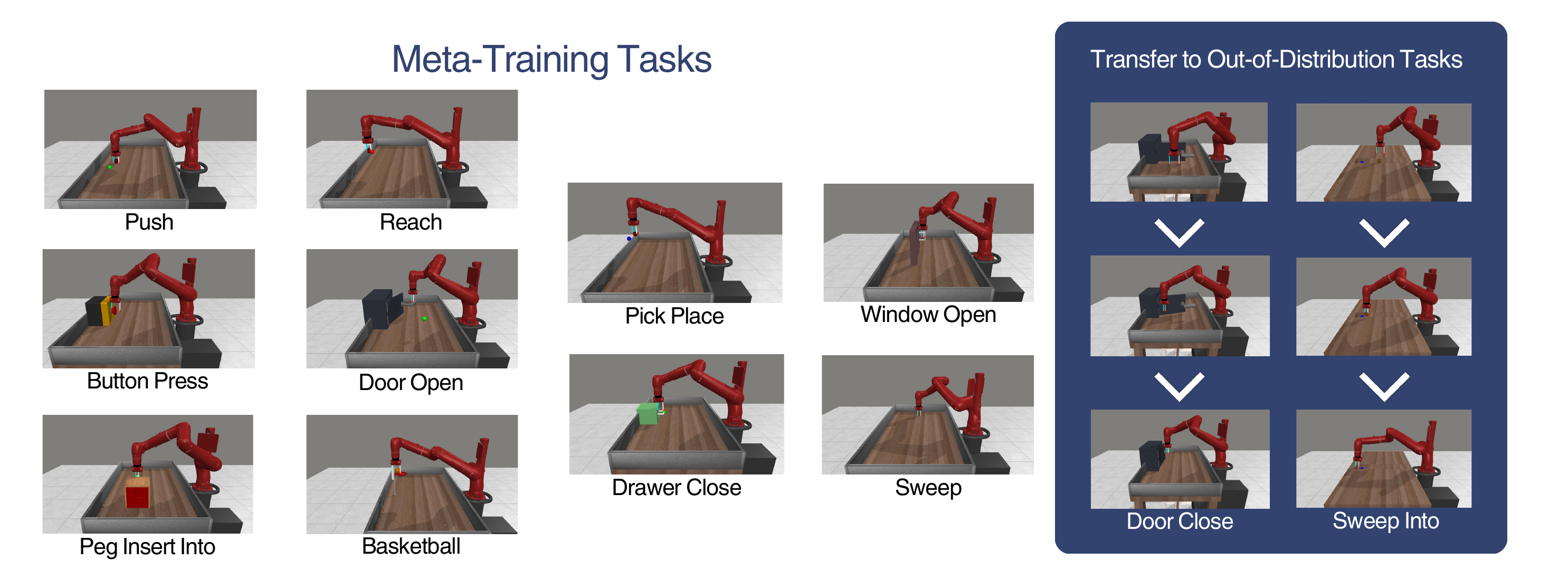}
    \caption{Illustration of the setting of the ML10 Benchmark in Meta-World. We evaluate HTrMRL in each task and collect these frames. On the left are the tasks HTrMRL is trained on, whereas on the right we show $3$ frames (\textit{beginning}, \textit{middle} and \textit{end} of the episode) from evaluating on $2$ of the $5$ test tasks. Though these tasks are not present in the training set, HTrMRL manages to generalize and successfully close the door and sweep the block into the hole.}
    \label{fig:plot_ml10_animation}
\end{figure*}
Meta-Reinforcement Learning~(Meta-RL)\footnote{For a recent survey we refer to \citet{beck2023survey}.} emerged as a framework aimed at making RL algorithms more sample-efficient by focusing on leveraging knowledge learned from previous tasks to speed up learning for the current task. Meta-RL optimizes a bilevel optimization problem during the \emph{meta-training} phase. The inner loop is usually represented by traditional RL algorithms and the outer loop represents the meta-learning algorithms, which exploit information on the inner loop to improve performance across multiple tasks. Once the meta-training phase is complete, only the inner loop algorithm is used in the next phase, called the \emph{adaptation} phase. Due to the meta-training phase, the inner loop algorithm can achieve good performance using fewer interactions with the new task compared to starting from scratch. \citet{pmlr-v70-finn17a} proposed a Model-Agnostic Meta-Learning~(MAML) algorithm, which can be used with any policy gradient algorithm as the inner loop to learn an initialization for the policy in order to speed up learning on a new test task.  \citet{Duan-rl2-16} proposed RL$^{2}$, an approach to learn a reinforcement learning algorithm using RNNs, which share the hidden state across episodes from the same task. RL$^{2}$ uses a policy gradient algorithm as the meta-learning algorithm in the outer loop. During the adaptation phase, learning occurs only within the dynamics of the RNN, without gradient adaptation.
Another approach proposed by \citet{pmlr-v97-rakelly19a}, PEARL, uses an off-policy RL algorithm, Soft Actor Critic~(SAC)~\citep{pmlr-v80-haarnoja18b}, as the outer loop, and in the inner loop, it augments the state with a stochastic latent representation of the task.
\citet{Zintgraf2020VariBAD} proposed VariBAD, a variant of RL$^{2}$ for Bayes-adaptive deep RL. VariBAD uses VAEs to learn an approximate posterior distribution over task features, which are used to augment the state as input to the policy. The outer loop algorithm for VariBAD is the Proximal Policy Optimization~\citep[PPO;][]{schulman2017proximal} algorithm, but it can be replaced by any policy gradient algorithm. During the adaptation phase, VariBAD uses only inference from the encoder of the VAE and the policy.

The transformer architecture~\citep{NIPS2017_vaswani} has revolutionized the field of Natural Language Processing~(NLP) and beyond. Relying on self-attention mechanisms enables it to process entire sequences of data simultaneously, a significant departure from the sequential processing of traditional RNNs and LSTMs. This parallel processing ability not only enhances efficiency, but also improves the handling of long-range dependencies in data.
\citet{pmlr-v119-parisotto20a} explored the application of transformer architecture in RL settings, finding that canonical transformers require certain adaptations for stable training in RL tasks. They addressed issues such as data efficiency and overfitting, contributing significantly to the understanding of transformers in RL contexts.
\citet{NEURIPS2021_chen_DecisionT} treat RL as a sequence modeling problem. This approach takes advantage of the scalability and simplicity of transformers to efficiently handle RL tasks. This approach differs from conventional RL methods that typically rely on the estimation of the value function or policy gradients.
\citet{NEURIPS2021_janner_offlineRL} extend this concept, showcasing how transformers can be used to unify various RL tasks into a single sequence modeling framework. This approach has shown promise in simplifying and improving offline RL algorithms.

Offline RL works use datasets of collected trajectories and a supervised prediction loss to train the transformer. This means that the learning process is based on a fixed dataset of experiences and that the model does not interact with the environment during training. Although effective for certain scenarios, this approach might not adapt well to dynamic or unpredictable environments where new experiences differ significantly from the training data.
\citet{lin-fmdm22} propose a transformer-based offline meta-RL approach to learn adaptable policies. Their contextual meta transformer aims at learning context embeddings based on episodic experiences to prompt a policy in a one-shot setting.
\citet{pmlr-v162-melo22a} explored the application of transformers in meta-reinforcement learning, demonstrating their potential to mimic memory reinstatement mechanisms. They proposed using a transformer architecture as the inner loop in RL$^{2}$. The transformer captures information from a sequence of the most recent $5$ transitions that are used as input. To predict the next action, the policy is conditioned solely on the output generated by the transformer. Similarly to \citet{pmlr-v162-melo22a}, we model meta-features directly into the policy network through a transformer policy.
In contrast, however, we propose a parameter-efficient hierarchical transformer architecture to take advantage of the intra-episode and inter-episode experience. We show that this design is more adept at capturing complex, multilevel patterns in the data, which is beneficial in diverse and dynamic task environments.

%% file: 3_preliminaries.tex
\label{sec:preliminaries}
The Reinforcement Learning (RL) problem is typically formalized within the framework of Markov Decision Processes (MDPs), defined by a tuple $(S, A, P, R, \gamma)$. $S$ represents a finite set of states in the task. Each state $s \in S$ encapsulates all the information that describes the situation at a particular point in time. $A$ denotes a set of actions that can be executed in all states. 
$P\colon S\times A\times S\to\left[0,1\right]$ is the state transition probability function that describes the dynamics of the MDP. More specifically, $P(s'|s, a)$ gives the probability of transitioning to state $s'$ when action $a$ is taken in state $s$. $R\colon S\times A\to\mathbb{R}$ is the reward function, with $R(s, a)$ giving the immediate reward received after taking action $a$ from state $s$. $\gamma\in\left[0,1\right]$ is the discount factor that determines the difference in importance between future rewards and current rewards. For an agent interacting with the MDP, lower values of $\gamma$ mean that the agent will prioritize immediate rewards more strongly, while a higher value indicates a preference for long-term gains.

The goal in RL is to find an optimal policy $\pi^*\in\Pi$, with $\pi\colon S\to A$, that maximizes the expected discounted cumulative reward. This can be formally defined as $\pi^* \in \argmax_{\pi\in\Pi} J(\pi)$ with
\begin{equation*}
     J(\pi)=\mathbb{E} \left[ \sum_{t=0}^{\infty} \gamma^t R(s_t, a_t) \mid s_0 = s, \pi \right]
\end{equation*}

Here, $J(\pi)$ denotes the expected cumulative reward when starting in $s_0 = s$, playing action $a_t$ in state $s_t$ as dictated by policy $\pi$, i.e., $a_t\sim\pi(s_t)$. Most modern RL approaches parameterize the policy as $\pi_\theta$, where the parameters $\theta$ are typically the weights of a neural network, and aim to find the parameters that satisfy the previous expression.

\textbf{Meta-Reinforcement Learning (Meta-RL):} 
Given a distribution of tasks $p(\mathcal{T})$, where each task $\tau \sim p(\mathcal{T})$ can be considered a distinct MDP with its own state space $S^\tau$, action space $A^\tau$, transition dynamics $P^\tau$ and reward function $R^\tau$, the objective of a Meta-RL technique is to learn a reinforcement learning algorithm $\mathcal{L}_\phi$ that directly outputs the parameters $\theta$ of a reinforcement learning policy.
The learned algorithm $\mathcal{L}$ is parameterized by $\phi$ such that $\mathcal{L}_{\phi} : \mathcal{T} \rightarrow \theta$ and aims to learn the parameters $\theta$ so that $\pi_\theta$ can generalize across a set of tasks sampled from the distribution $p(\mathcal{T})$ .
To this end, the policy interacts with a series of tasks $\tau$, drawn from $p(\mathcal{T})$, and aims to maximize its performance across this task distribution. This process involves two key phases:

\textbf{\emph{Meta-Training and Meta-Testing Phases:}}
During meta-training, the agent learns a policy $\pi_{\theta}$ and an update rule $\mathcal{L}_{\phi}$ (parameterized by $\theta$ and $\phi$, respectively) that can quickly adapt to new tasks, resulting in a bilevel optimization problem. This further involves learning a generalizable representation or prior knowledge that is transferable across tasks. The outer objective in this phase is to optimize the parameters $\phi$ such that $\phi^*\in \argmax_\phi G(\phi)$ with $G(\phi) = \mathbb{E}_{t\sim p(\mathcal{T})}\left[J(\pi_\theta)|\mathcal{L_\phi}, t\right]$ whereas the inner objective involves finding the weights of the policy that maximize $\theta^* \in \argmax_\theta J(\pi_\theta)$. In the meta-testing (adaptation) phase, the agent encounters new tasks $\tau_{\text{new}}$ sampled from $p(\mathcal{T})$. Using $\pi_{\theta^*}$ and $\mathcal{L}_{\phi^*}$, the agent quickly adapts to each new task. The adaptation effectiveness is measured by the agent's performance on these new tasks after a limited number of learning steps or experiences.

%
%


%% file: 4_method.tex
%
\begin{figure*}[ht!]%
    \centering%
    \includegraphics[width=\textwidth]{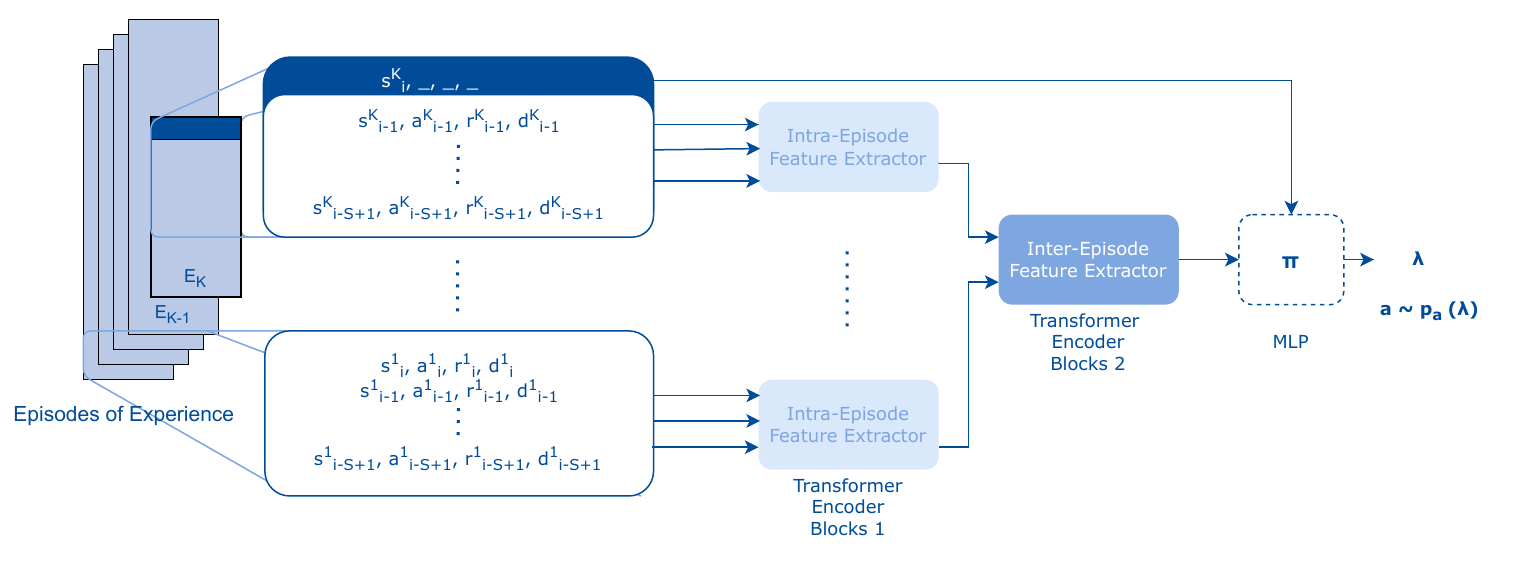}%
    \vspace{-1cm}
    \caption{Illustration of our HTrMRL architecture. We store the $K$ past episodes~(E) and sample a sequence of transition sequences of length $S$ from each episode, including a sequence of the current episode E$_{K}$, which ends with the current state. Transition sequences store which state $s_t$, action $a_t$, and reward $r_t$ were observed, but also a termination flag $d_t$ that indicates if an episode already terminated at time $t$. By inputting the sequences independently through a group of Transformer Encoder Blocks, we generate a feature vector for each sequence. We then put these sequence representations through a second group of Transformer Encoder Blocks to generate feature vectors that augment the state as input to policy $\pi$.}%
    \label{fig:htmrl_diagram}%
\end{figure*}%
\section{Deeper Memory Mechanisms in Transformer-Based Meta-RL}\label{sec:method}
We begin this section, with a critique of the limitations of current transformer-based Meta-RL approaches, particularly their narrow focus on intra-episodic memory. We introduce an enhanced architecture that broadens memory capabilities and facilitates deeper understanding between episodes. Finally, we discuss the architecture's memory complexity, underscoring its efficiency for use in Meta-RL.
\subsection{The Intra-Episodic Memory Hurdle}
It was recently demonstrated that transformers are capable of meta-reinforcement learning \citep{pmlr-v162-melo22a}. To this end, the TrMRL agent architecture extends the RL$^2$ \citep{Duan-rl2-16} framework for online Meta-RL.
In RL$^2$ an RL algorithm is encoded by an RNN (see \cref{fig:rl2_framework} in \cref{app:rl2_framework}) where the outer loop parameters are given by the RNN weights and the inner loop dynamics is encoded by the hidden state of the RNN.
TrMRL builds on this approach and, crucially, however, replaces the RNN by a transformer architecture.
Furthermore, \citet{pmlr-v162-melo22a} showed that the self-attention mechanism used in the proposed transformer enables a form of episodic memory. This, in turn, facilitates easier identification of the underlying MDP, facilitating better learning dynamics of the policy by contextualizing the agents' behavior with respect to the relevant task dynamics.
However, despite TrMRL's success in generating adaptable policies for tasks with similar structures, it faces challenges in more diverse or heterogeneous settings. To address this limitation, we suggest improving the memory component of the model. Our proposed improvement not only preserves information within individual episodes, but also across a series of episodes, fostering a broader understanding and generalization across different and diverse tasks.

\begin{figure*}[ht]%
    \centering%
%
    \begin{minipage}{0.49\textwidth}%
        \centering%
        \includegraphics[width=1.2\textwidth]{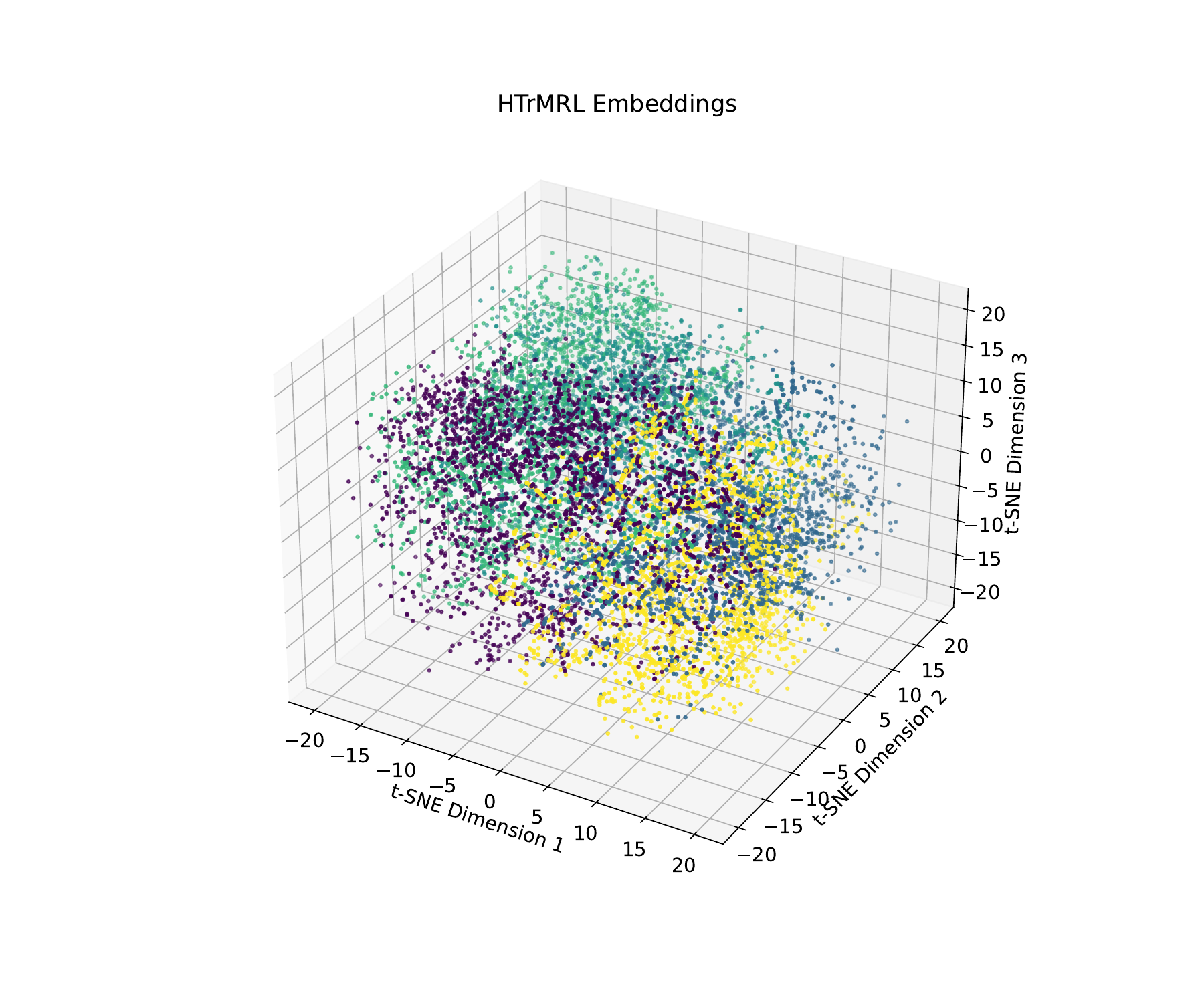}%
    \end{minipage}%
    \begin{minipage}{0.49\textwidth}%
        \centering%
        \hspace{-2em}%
        \includegraphics[width=1.2\textwidth]{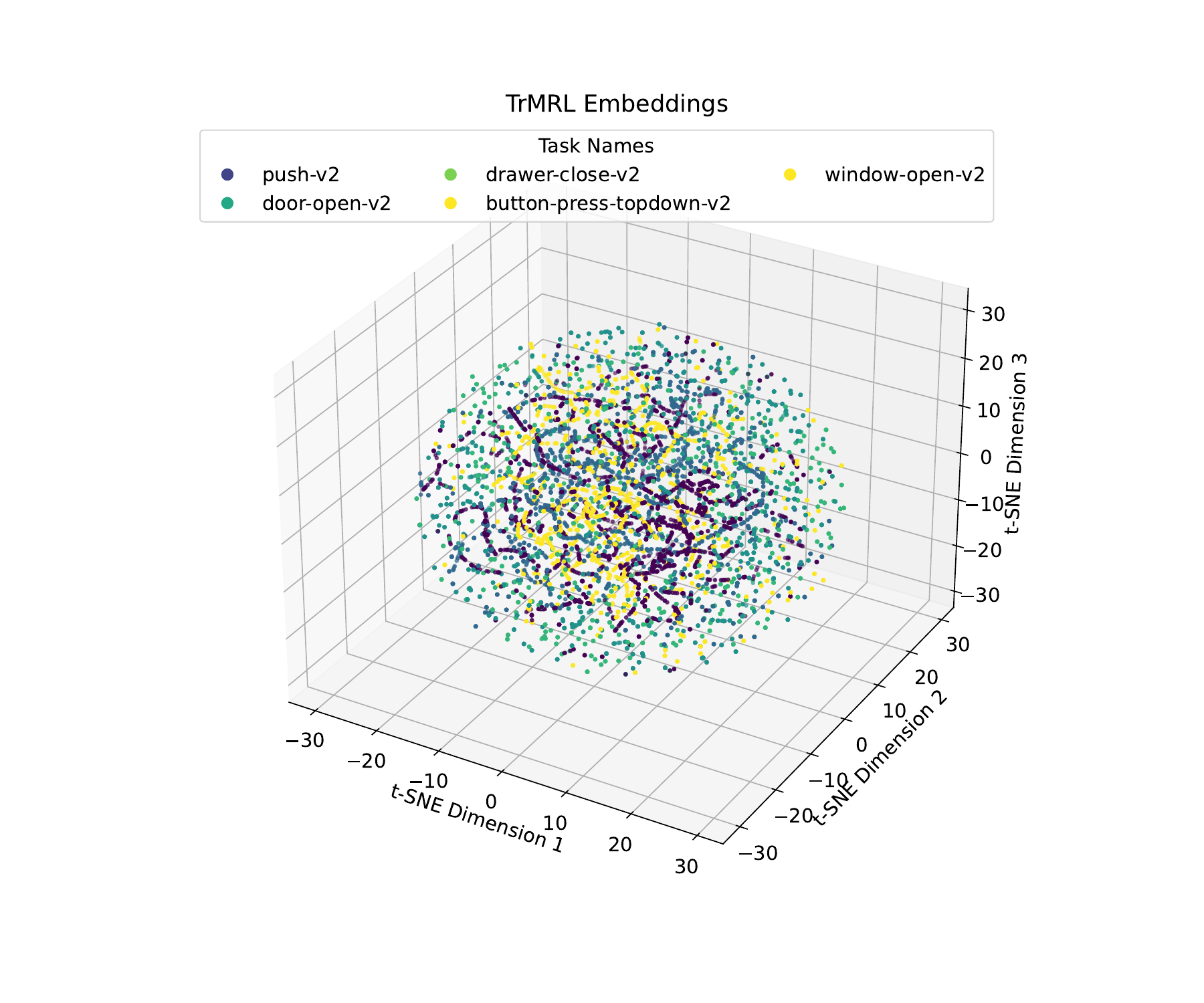}%
    \end{minipage}%
    \vspace{-2.75em}%
    \caption{T-SNE plots of the output embeddings for our HTrMRL(left) and TrMRL(right) for five of the tasks in the training set of the ML10 benchmark of Meta-World. For better visibility, we only plot $5$ tasks here but provide a version with all $10$ in \protect\cref{fig:TSNE_plots_all}.}%
    \label{fig:TSNE_plots}%
\end{figure*}

\subsection{Hierarchical Transformers for Meta-RL}
We illustrate our proposed approach, which we dub HTrMRL (\textbf{H}ierarchical \textbf{T}ransformers for \textbf{M}eta-\textbf{R}einforcement \textbf{L}earning), in Figure~\ref{fig:htmrl_diagram}. Our hierarchical architecture consists of two distinct groups of transformer encoder blocks.
The first set of encoder blocks (denoted as $\mathbf{T}_{1}$) encode the same intra-episodic memory as proposed in TrMRL and allow us to understand the transition dynamics within episodes. The second set of encoder blocks ($\mathbf{T}_{2}$) accesses the intra-episodic memories to encode how the transition dynamics across episodes relate to each other. In order to enable learning with this architecture, we store K episodes including the most recent episode and sample one sequence of length S from each episode. For the current episode, we ensure that the sequence ends with the current state observation of the agent such that it can choose the most appropriate next action. The K sequences of transitions of length S are stored as tuples
$\chi^k_i=(s^k_i, a^k_i, r^k_i, d^k_i)$
where $d$ is the termination flag, $0\leq i < S$ is the index of the transition within the sequence, and $k$ is the index of the episode from which we sample the sequence.

The intra-episode feature extractor, $\mathbf{T}_{1}$, aims to learn a representation that describes the intra-episode dynamics from the transitions that serve as input. The blocks $\mathbf{T}_{1}$ do the following transformation for each of the sequences sampled:
\begin{align*}
    \mathbf{z}_{seq}^{k}\;=&\;\mathbf{T}_{1}\left(\varphi_{1}(\chi^k_0), \varphi_{1}(\chi^k_1), \ldots, \varphi_{1}(\chi^k_S) \right)
\end{align*}

where $\varphi_{1}$ is a linear transformation we apply to the transitions, $k$ is the index of the episode from which the sequence was sampled, $k \in \{1, ..., K\}$, and $\mathbf{z}_{seq}^{k}$ is the output for the last transition of the sequence. As the ordering of the transitions in the sequence is important information regarding the intra-episode dynamics, we use $\mathbf{T}_{1}$ with positional encoding on its inputs.

We then input intra-episode features $\mathbf{z}_{seq}^{k}$ from different episodes into the inter-episode feature extractor, $\mathbf{T}_{2}$, which learns to output inter-episode features as the representation of the task. The $\mathbf{T}_{2}$ blocks do the following transformation for the input:
\begin{equation*}
    \mathbf{z}_{task} = \mathbf{T}_{2}(\mathbf{z}_{seq}^{1},... ,\mathbf{z}_{seq}^{K-1}, \mathbf{z}_{seq}^{K}) 
\end{equation*}
where $\mathbf{z}_{task}$ is the output for the last input $\mathbf{z}_{seq}^{K}$. We consider that the ordering of $\mathbf{z}_{seq}$ is not important for learning useful inter-episode features. That is why we use $\mathbf{T}_{2}$ without positional encoding in its inputs. 
We condition the policy on the learned inter-episode features and a linear transformation $\varphi_{2}$ of the current state $s$: 
\begin{equation*}
    \mathbf{\lambda} = \pi(\varphi_{2}(s), \mathbf{z}_{task})
\end{equation*}
where the output of the policy, $\mathbf{\lambda}$, represents the parameters of the distribution $p_{a}$ from which we sample the next action, i.e., $a \sim p_{a}(\mathbf{\lambda})$. We use a Gaussian distribution as the model for the probability distribution of the actions, thus, specifically, $\mathbf{\lambda}$ is the mean and variance of the distribution. 

With the intention of capturing more global characteristics of the task, we focus on sampling K sequences of transitions from different episodes of experience, where S represents the number of transitions in a sequence. Using only one group of transformer encoder blocks to process this data (i.e. E$\cdot$S transitions) requires an attention matrix of complexity $O(S^{2}K^{2})$. On the other hand, using one transformer encoder block to capture intra-sequence features $\mathbf{z}_{seq}$ requires an attention matrix of complexity $O(S^{2})$. Capturing inter-sequence features $\mathbf{z}_{task}$ from the output of the sequence transformer encoder block requires an attention matrix of complexity $O(K^{2})$. Thus, a hierarchical architecture of the transformer is a parameter-efficient design that scales better ($O(S^{2}+K^{2})$) to longer sequences and more episodes than a non-hierarchical one in orders of magnitude. 

%% file: 5_research_questions_results.tex
\begin{figure*}[ht!]
    \centering
    \includegraphics[width=0.95\textwidth]{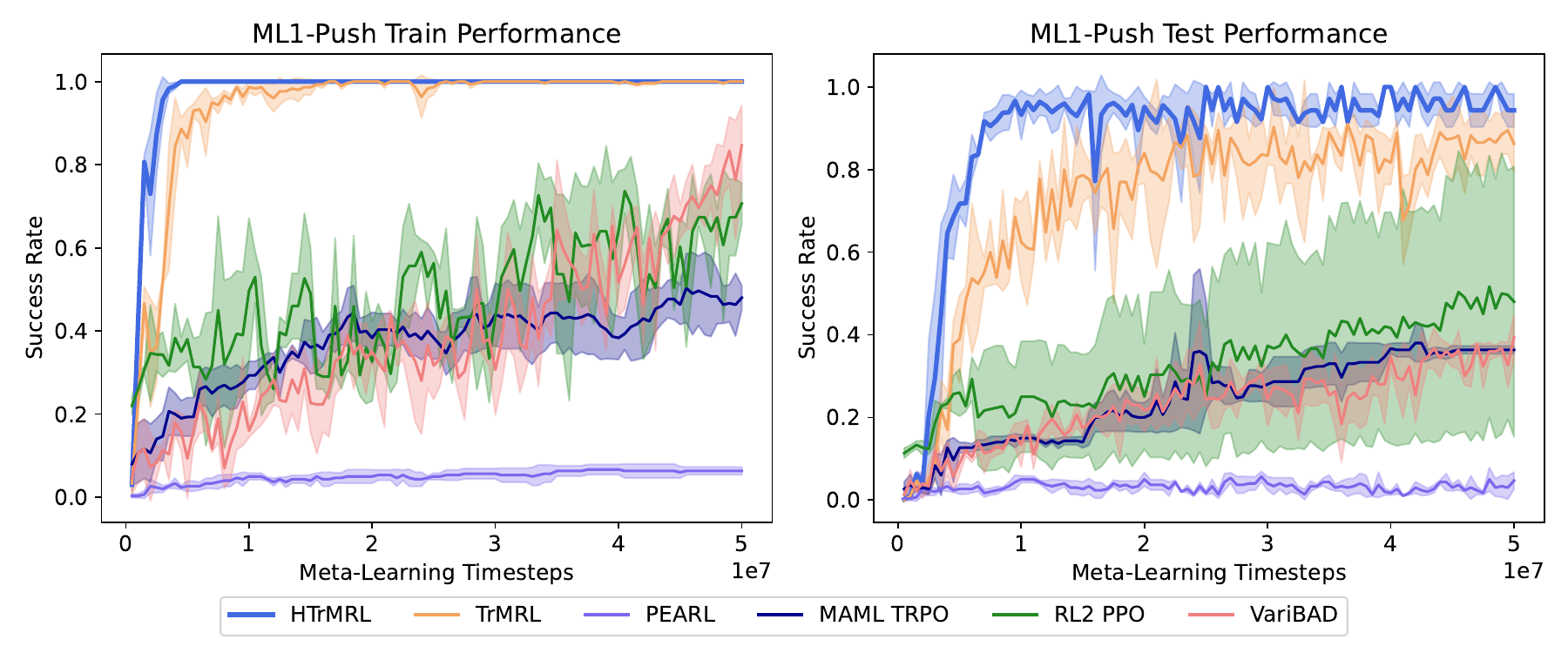}
    \caption{Meta-Train and Test performance in terms of Average Success Rate of HTrMRL, TrMRL, PEARL, MAML TRPO, RL2 PPO and VariBAD on the ML1 benchmark for training(left) and testing(right) on parametric variations of the \textit{Push} task.}
    \label{fig:plot_ml1_push_average_success}
\end{figure*}

\begin{figure*}[ht!]
    \centering
    \includegraphics[width=0.95\textwidth]{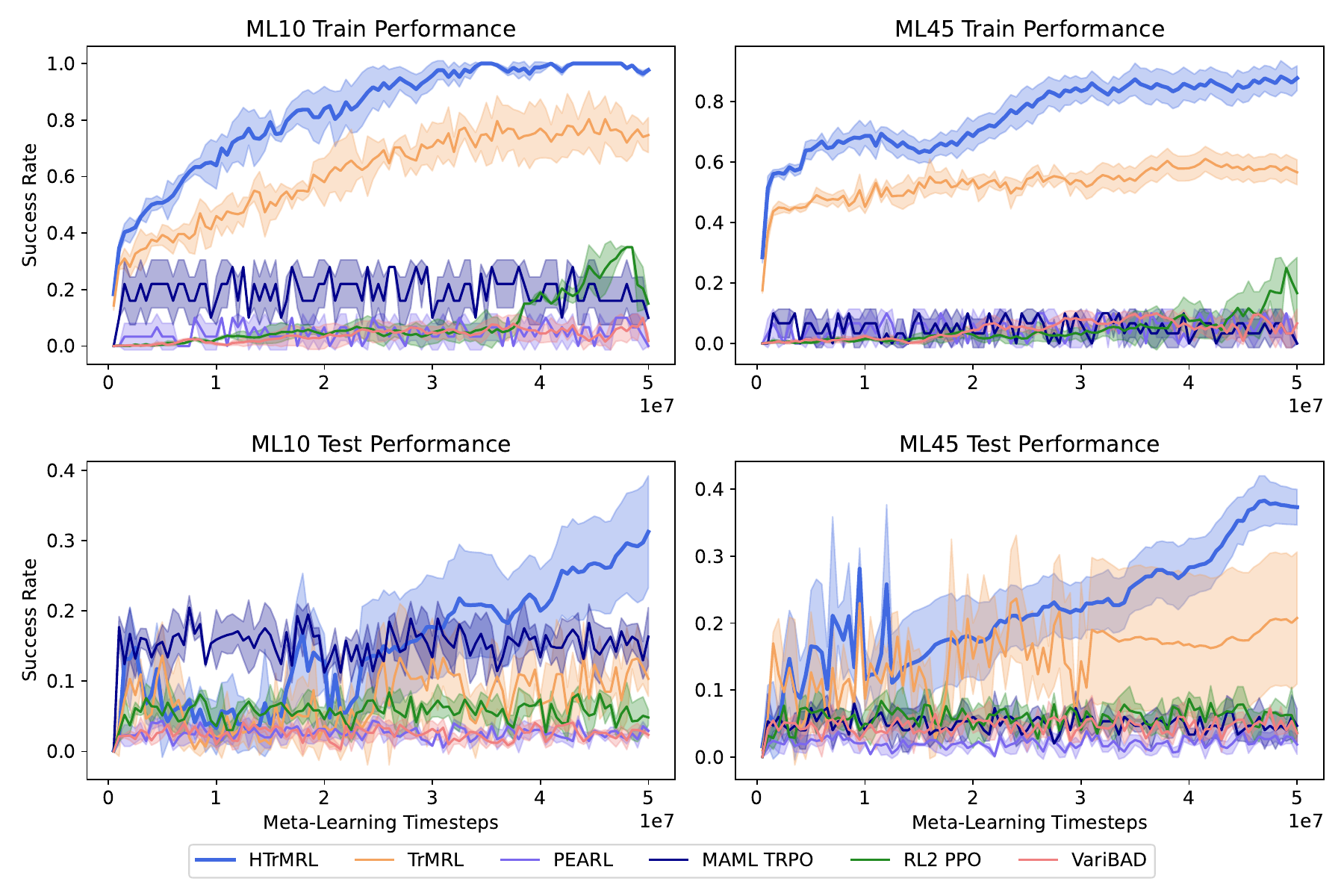}
    \caption{Meta-Train and Test performance in terms of Average Success Rate of HTrMRL, TrMRL, PEARL, MAML TRPO, RL2 PPO and VariBAD on the ML10(left) and ML45(right) benchmarks. Here, we are training and testing on disjoint sets of tasks.}\label{fig:plot_ml10_ml45_average_success}%
\end{figure*}%
\begin{figure*}[ht!]
    \centering
    \includegraphics[width=0.95\textwidth]{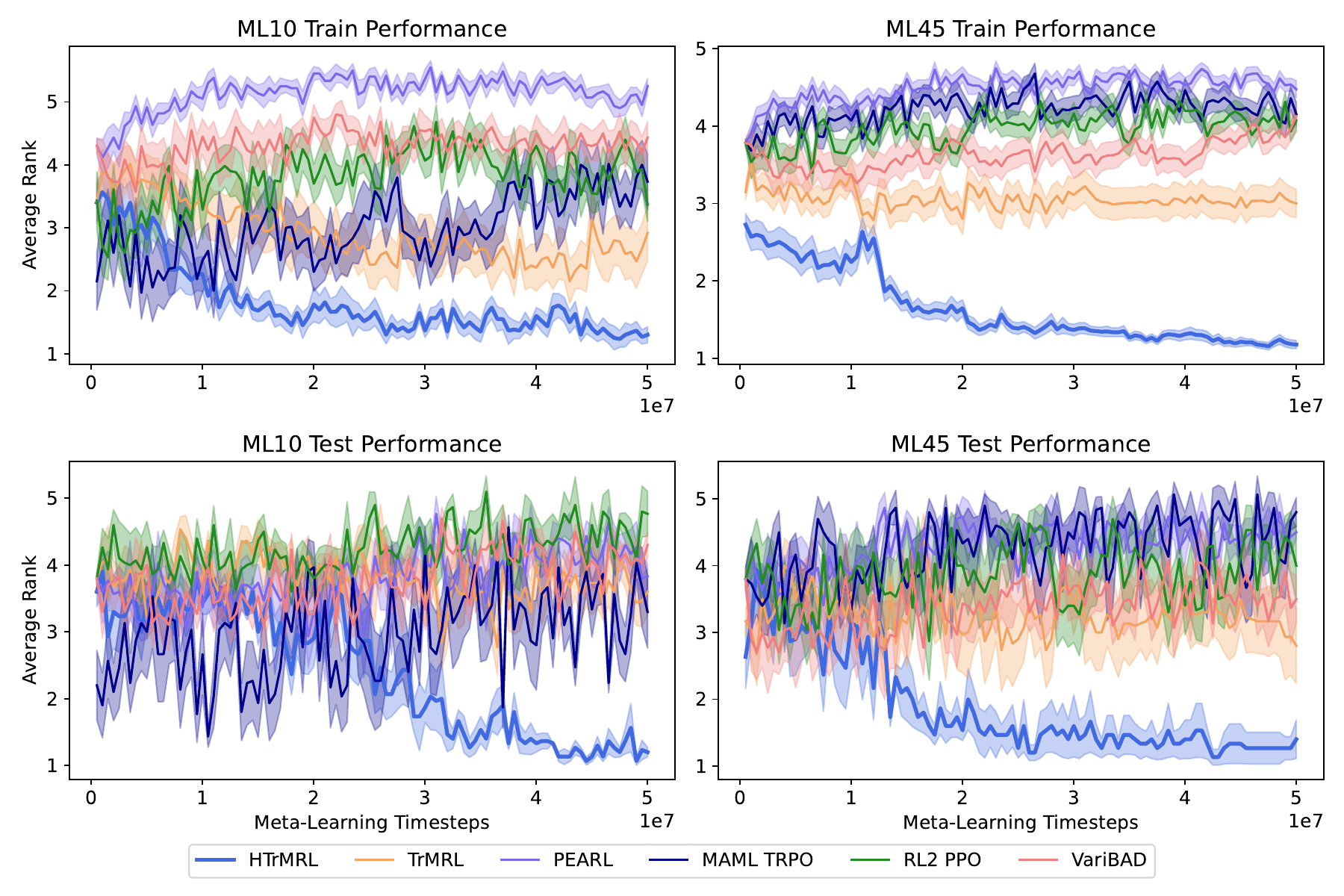}
    \caption{Average ranking plots of Meta-Train and Test performance for all methods on the ML10(left) and ML45(right) with disjoint train and test tasks. Lower ranks indicate better performance. The corresponding success rate plot can be found in \protect\cref{fig:plot_ml10_ml45_average_success}.}
    \label{fig:plot_ml10_ml45_ranks}
\end{figure*}
\label{sec:research_questions_results}
\subsection{Experimental Protocol}
\paragraph{Meta-World} To assess the performance of our hierarchical transformer approach, we use the Meta-World Benchmark~\citep{yu2019meta}. This benchmark is designed to help evaluate the performance of Meta-RL baselines in $50$ different simulated robotic manipulation tasks. These tasks all share the same state $S$ and action space $A$ as they are all located on a tabletop environment with a Sawyer arm (see \cref{appendix:envs} for details and illustrations of all tasks). What differs in each task is the behavior required to complete it successfully. For meta-learning, Meta-World includes three modes with varying degrees of difficulty (for details, see \cref{appendix:envs}):
\begin{itemize}\setlength\itemsep{.1em}
    \item \textbf{ML1} is designed to evaluate few-shot adaptation to parametric variations within one task.
    \item \textbf{ML10} is designed to evaluate few-shot adaptation to $5$ unseen test tasks, after training on $10$ tasks.
    \item \textbf{ML45} is similar to, but more complex than ML10 with $45$ training and $5$ test tasks.
\end{itemize}
For ML1, the desired goal position is not provided as input, so the meta-RL algorithm needs to determine the location of the goal through trial and error. Similarly, for ML10 and ML45, task IDs are not provided as input, so the meta-RL algorithm needs to identify tasks from experience.

\paragraph{Performance metrics} The authors of Meta-World propose the \textit{success rate} as a performance evaluation metric. The distance of the task-relevant object from the goal position determines whether the agent is successful or not in an episode. Thus, the success rate is the fraction of episodes where the agent manages to place the task-relevant object inside a pre-set perimeter around the goal position.
In the following we report averaged results for all Meta-World settings, but we provide the task-specific success-rate plots in \cref{app:additional_results}.

Due to the fact that the tasks in Meta-World have different levels of difficulty, we additionally use the \textit{average rank} as a performance metric for aggregating the performance across different tasks. We calculate the rank of methods over time for each task in the set based on their success rate, and then average over those ranks across tasks. 

\textbf{Baselines:}
We focus on investigating the performance of HTrMRL compared to other online meta-reinforcement learning methods, namely MAML~PPO~\citep{pmlr-v70-finn17a}, RL$^{2}$~\citep{Duan-rl2-16}, PEARL~\citep{pmlr-v97-rakelly19a}, VariBAD~\citep{Zintgraf2020VariBAD} and TrMRL~\citep{pmlr-v162-melo22a}. Detailed descriptions of each are given in \cref{appendix:baselines}.
We trained all methods for $5\times10^7$ steps for $5$ seeds. When comparing the methods in  the figures, we plot the mean and standard error of the performance metrics across seeds. 

%
%
%
%

\subsection{Hypotheses and Results}
\paragraph{\emph{Hypothesis 1:} HTrMRL captures more general task features through its ability to capture intra- as well as inter-episode experiences and outperforms baselines that only use intra-episode experiences.}

\input{ablations}
\textbf{Results:} We compare the embeddings that HTrMRL learns to those of TrMRL to investigate the extent to which they differentiate between tasks. We sample episodes from the training tasks in ML10 and plot the output of the transformers for both HTrMRL and TrMRL through dimensionality reduction. For the sake of visual clarity, in Figure~\ref{fig:TSNE_plots} we show the results for $5$ of these tasks. For the plot with all the $10$ tasks, see Figure~\ref{fig:TSNE_plots_all} in \cref{app:additional_results}. The learned embeddings of HTrMRL produce a clearer grouping of the embeddings for the same task compared to the embeddings of TrMRL. Furthermore, we can see a grouping of tasks that are similar to each other, e.g., \textit{Window Open} and \textit{Door Open}. Through learning from inter-episode experiences, HTrMRL can better differentiate between the different tasks in the ML10 training set. In contrast, TrMRL only focuses on the intra-episode experiences, which, in this case, do not provide sufficient information to differentiate between the diverse tasks. For easier analysis, we provide scripts for interactive versions of the plots in our provided code repository.

\textbf{\emph{Hypothesis 2:} HTrMRL outperforms the state-of-the-art in online meta-RL algorithms in few-shot adaptation to parametric variations of tasks.}
\textbf{Results:} We compare HTrMRL and the baselines in the ML1 benchmark to evaluate adaptability to parametric variations of the \textit{Push} and \textit{Reach} tasks in Meta-World. We show the results in terms of Success Rate in Figure~\ref{fig:plot_ml1_push_average_success} for \textit{Push}. The figure contains results for meta-training (left) and test (right) performance in the case where we train on variations of the \textit{Push} task, and test the adaptability of the methods on a set of unseen variations of the same task. In terms of meta-training performance (left), we can see that transformer-based methods are faster in differentiating between the variations of the task and thus achieve a maximum success rate of $1.0$. 
In terms of test performance (right), HTrMRL and TrMRL generalize successfully, reaching a test success rate greater than $0.8$. In addition, HTrMRL shows a higher generalizability compared to TrMRL. RL2 PPO struggles to robustly generalize, exhibiting a higher variation in performance compared to the other methods.
In Figure~\ref{fig:plot_ml1_reach_average_success} in \cref{app:additional_results} we show the Meta-Train and Test performance on variations of the \textit{Reach} task. The task variability for \textit{Reach} is not as difficult to learn, 
 thus all the methods perform better both when meta-training and testing. 
\paragraph{\emph{Hypothesis 3:} 
HTrMRL outperforms the state-of-the-art online meta-RL in out-of-distribution~(OOD) tasks.}
\textbf{Results:} We define out-of-distribution~(OOD) performance to be the performance of a meta-learned agent on a set of tasks that is disjoint from the tasks it is meta-learned on. 

We show the performance of the methods in terms of the average success rate for the meta-training~(top) and test~(bottom) performance for ML10~(left) and ML45~(right) in Figure~\ref{fig:plot_ml10_ml45_average_success}. HTrMRL is the only method that achieves an average success rate greater than $0.8$ in ML10 and greater than $0.6$ for ML45 in terms of meta-training performance with a final success rate roughly $20\%$ higher than the baselines. The other methods fail to learn policies that successfully identify the different tasks in the training set for the given budget of $5\times10^{7}$ meta-learning timesteps, and thus struggle to solve them. In terms of OOD performance, the bottom row shows that generalization to distinct tasks is still a challenge for meta-RL. HTrMRL achieves an average test success rate of approximately $0.35$ for ML10 and $0.4$ for ML45. 
The difference in test performance for HTrMRL and TrMRL further demonstrates that inter-episode features can improve meta-RL performance without incurring the cost of gradient adaptation at test time.

 
 To focus on comparing the methods based on their robustness in relative performance to each other, we further provide plots showing the average ranks of the methods as a performance metric. We show these plots in Figure~\ref{fig:plot_ml10_ml45_ranks} in the main paper and Figure~\ref{fig:plot_ml1_ranks} in \cref{app:additional_results}. These plots show that HTrMRL robustly achieves the highest relative performance (and thus the lowest rank) across the four meta-learning benchmarks we consider.
For a detailed illustration of performance in terms of the success rate per task, we refer to \cref{fig:plot_ml10_average_success_rate_train_task} -~\ref{fig:plot_ml45_average_success_rate_test_task} in \cref{app:additional_results}. Furthermore, we present the performance in terms of average ranks per task in Figures~\ref{fig:plot_ml10_rank_train_task} -~\ref{fig:plot_ml45_rank_test_task} in \cref{app:additional_results}. 
\subsection{Ablations}
To investigate the behavior of HTrMRL with respect to the number of episodes (K) sampled and the sequence length (S), we provide ablation results in \cref{tab:htrmrl_performance}. We also ablate the sampling strategy (sampling one transition sequence per episode vs. sampling multiple sequences randomly from any episode), design choices (concatenating a linear transformation of the state as input to the policy $\pi$ vs using only the transformer output as the input to the $\pi$, as TrMRL does; and the number of encoder blocks in the architecture). The configuration we use in our experiments is K=$25$, S=$5$, and we use two encoder blocks for $\mathbf{T}_{1}$ and $\mathbf{T}_{2}$ each. For a detailed comparison in terms of performance over meta-learning timesteps, we refer to \cref{fig:plot_ml1_e_and_s_ablation,fig:plot_ml1_r_s_and_a_ablation} in \cref{app:additional_results}.

%% file: ablations.tex
\begin{table*}[hbpt]
\centering
\small
\begin{tabular}{c|cccccc}
\toprule
\backslashbox{\textbf{S}}{\textbf{K}} & \textbf{5} & \textbf{10} & \textbf{15} & \textbf{20} & \textbf{25} & \textbf{50} \\
\midrule
\textbf{5} & 0.997 $\pm$ 0.005 & 1.000 $\pm$ 0.000 & 0.997 $\pm$ 0.005 & 1.000 $\pm$ 0.000 & 1.000 $\pm$ 0.000 & 0.909 $\pm$ 0.058 \\
\textbf{25} & 0.997 $\pm$ 0.005 & 1.000 $\pm$ 0.000 & 1.000 $\pm$ 0.000 & 0.993 $\pm$ 0.009 & 0.924 $\pm$ 0.053 & 1.000 $\pm$ 0.000 \\
\textbf{50} & 0.963 $\pm$ 0.039 & 0.960 $\pm$ 0.050 & 0.993 $\pm$ 0.005 & 0.827 $\pm$ 0.098 & 0.990 $\pm$ 0.014 & 0.893 $\pm$ 0.041 \\
\bottomrule
\end{tabular}
\caption{HTrMRLs final meta-training success-rate on \textbf{ML1 Push} when varying the number of episodes (K) and the \textbf{S}equence length.}
\label{tab:htrmrl_performance}
\end{table*}

%% file: 6_conclusion.tex
We introduced Hierarchical Transformers for Meta-Reinforcement Learning (HTrMRL), a powerful approach to online meta-RL. Our experiment within the Meta-World benchmark highlights our method's capability for few-shot adaptation to tasks with parametric variations, as well as its capacity to adjust to completely unseen tasks that share the same state and action spaces as the training tasks. The results demonstrate that HTrMRL outperforms existing state-of-the-art online meta-RL methods by leveraging inter-episodic features, particularly in scenarios requiring rapid adaptation to a variety of tasks. Despite these advances, the challenge of adapting to new tasks efficiently persists. 
This underscores the importance of learning to compress the agent's experiences with inter-episodic features marking a significant step forward toward meta-RL methods with better adaptation capabilities. 

%% file: 7_appendix.tex
\subsection{RL$^{2}$ Framework}\label{app:rl2_framework}
\begin{figure*}[ht!]
    \centering
    \includegraphics[width=\textwidth]{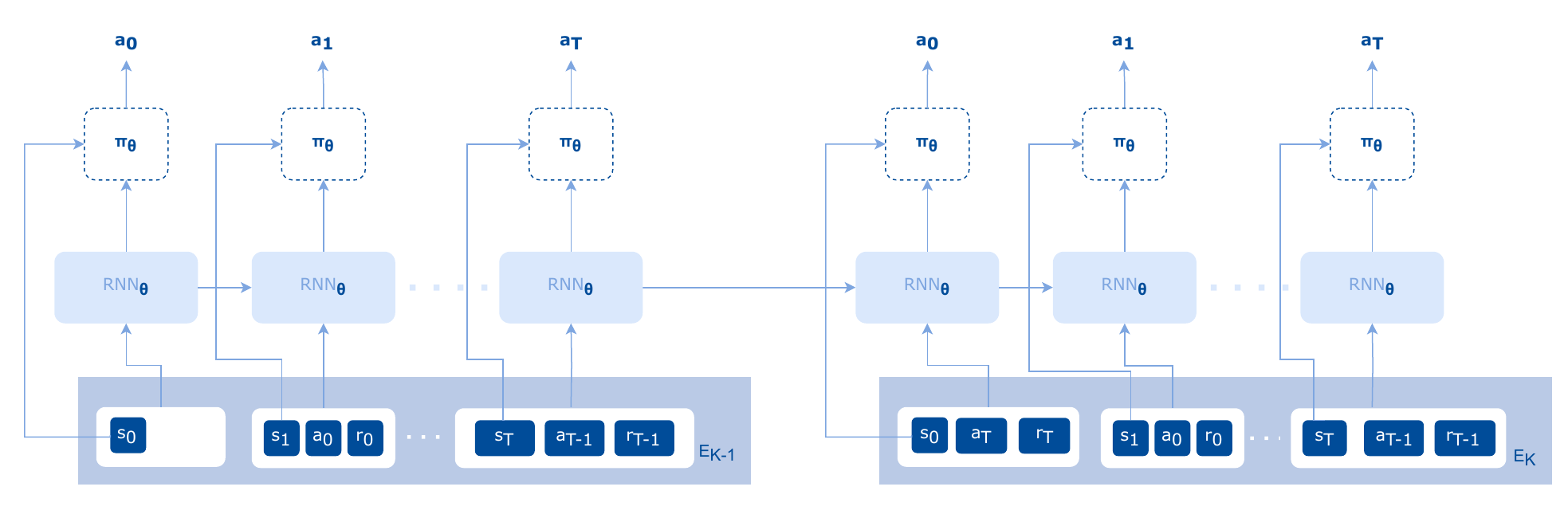}
    \caption{Illustration of the RL$^{2}$ framework, based on Figure 4 from \citet{beck2023survey}. As illustrated, the RNN takes as input the current state $s_{t}$, the action(if there is any) that lead to the state and the reward incurred from reaching that state. The output of the RNN is then used as input to the policy $\pi$ along with the current state.}
    \label{fig:rl2_framework}
\end{figure*}

\subsection{Benchmark Details}\label{appendix:envs}
\begin{figure*}[h!]
    \includegraphics[width=\textwidth]{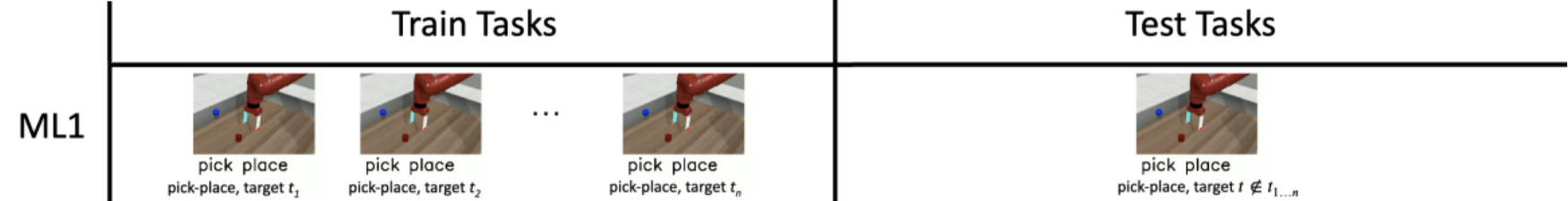}
    \caption{Meta-world Tasks in the M1 with training/test Split. Figure taken from \protect\citep{yu2019meta}}
    \label{fig:plot1-metaworld}
\end{figure*}
\paragraph{ML1} provides one main task, with parametric variation in object and goal positions. ML1 enables evaluation in the few-shot adaptation setting. In particular, adaptations to parametric variations within one task. \cref{fig:plot1-metaworld} gives an example of meta-training and testing for the pick in place task, where only the test-task target position is different from those observed during training.
\begin{figure*}[h!]
    \includegraphics[width=\textwidth]{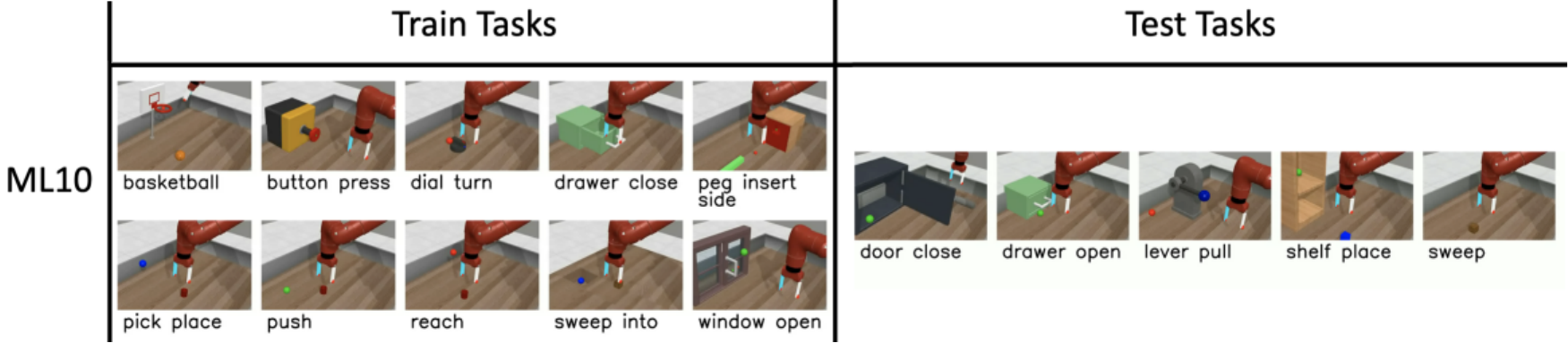}
    \caption{Meta-world Tasks in the M10 protocol with training/test Split. Figure taken from \protect\citep{yu2019meta}}
    \label{fig:plot2-metaworld}
\end{figure*}
\paragraph{ML10} provides $10$ different training tasks and $5$ others to test the adaptation performance of meta-learning methods. This mode evaluates few-shot adaptation to $5$ new test tasks, after training on $10$ different tasks, see \cref{fig:plot2-metaworld}. The test tasks are all related to the training tasks, but unseen operations have to be performed (such as opening a drawer at test time instead of closing it as learned in meta-training).
\begin{figure*}[h!]
    \includegraphics[width=\textwidth]{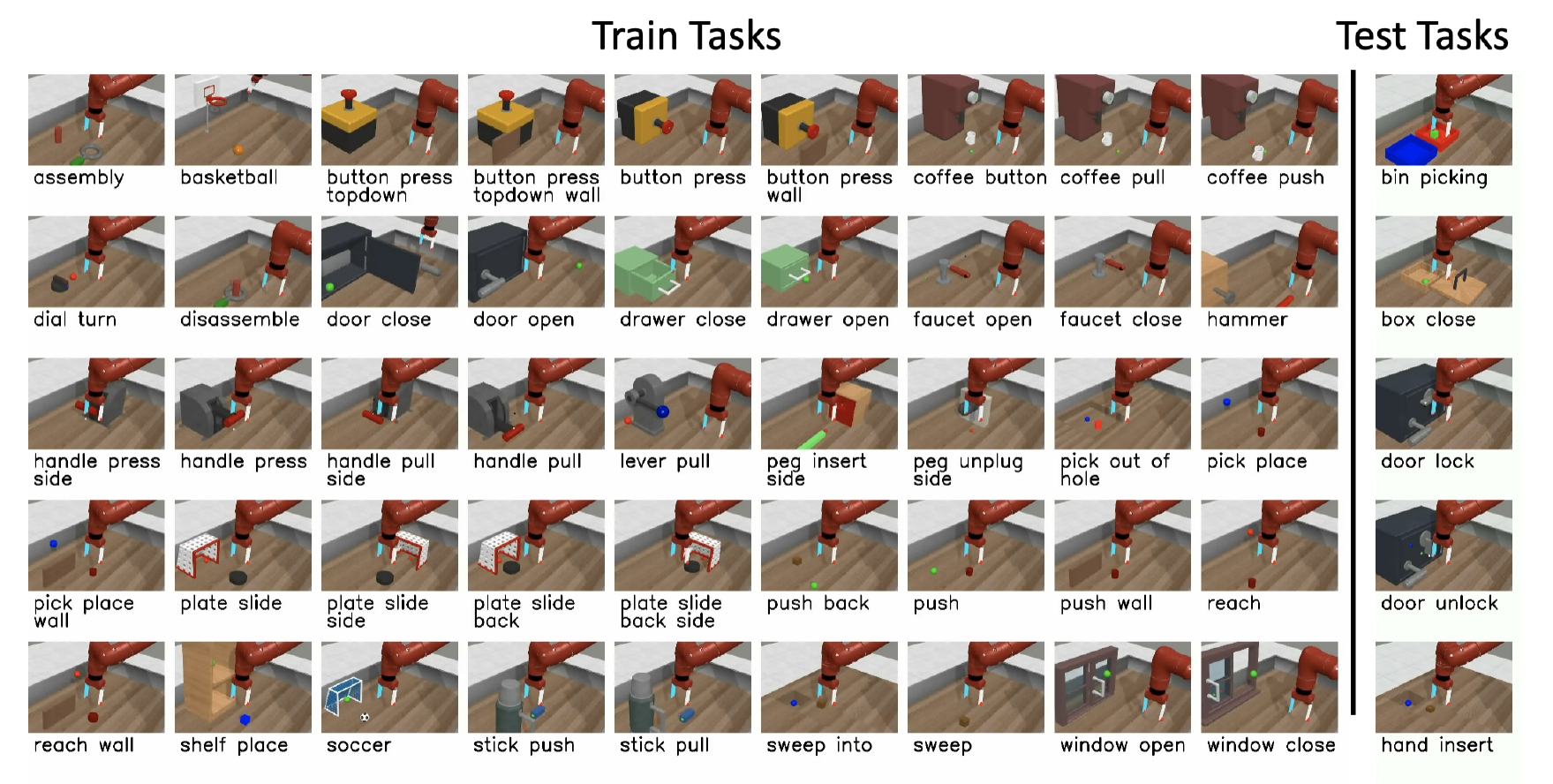}
    \caption{Meta-world Tasks in the M45 protocol with training/test Split. Figure taken from \protect\citep{yu2019meta}}
    \label{fig:plot3-metaworld}
\end{figure*}
\paragraph{ML45} provides $45$ highly different training tasks and $5$ others to test the adaptation performance of meta-learning methods. This mode evaluates few-shot adaptation to $5$ new test tasks, after training on $45$ different tasks. \cref{fig:plot3-metaworld} shows all $50$ tasks.

\subsection{Baselines}\label{appendix:baselines}
\textbf{MAML PPO} is the application of the MAML meta-learning algorithm for meta-RL~\citep{pmlr-v70-finn17a}. The inner-loop algorithm is PPO, whereas the outer-loop algorithm is the MAML algorithm. We use the implementation provided in the \textit{garage} repo~\cite{garage}.

\textbf{RL$^{2}$} uses an RNN in the inner loop as described in Section~\ref{sec:related_work}, and TRPO in the outer loop~\citep{Duan-rl2-16}. We use the implementation provided in the \textit{garage} repo~\cite{garage}, which replaces TRPO with PPO.

\textbf{PEARL} is an off-policy meta-RL algorithm, in contrast to the other methods we compare with~\citep{pmlr-v97-rakelly19a}. We use the implementation provided in the \textit{garage} repo~\cite{garage}.

\textbf{VariBAD} uses a VAE to learn a distribution of task features that increase the state space~\citep{Zintgraf2020VariBAD}. In contrast to PEARL, VariBAD uses PPO in the outer loop, which is an on-policy meta-RL algorithm. We use the implementation provided by the authors. 

\textbf{TrMRL} is the closest baseline to our proposed method~\citep{pmlr-v162-melo22a}. It uses a transformer in the inner loop of the RL$^{2}$ algorithm. It takes a sequence of the $5$ most recent transitions as input to the transformer. We use the implementation provided by the authors. 

\subsection{Additional Results}
\label{app:additional_results}

\begin{figure*}[htbp!]
    \centering
    \includegraphics[width=\textwidth]{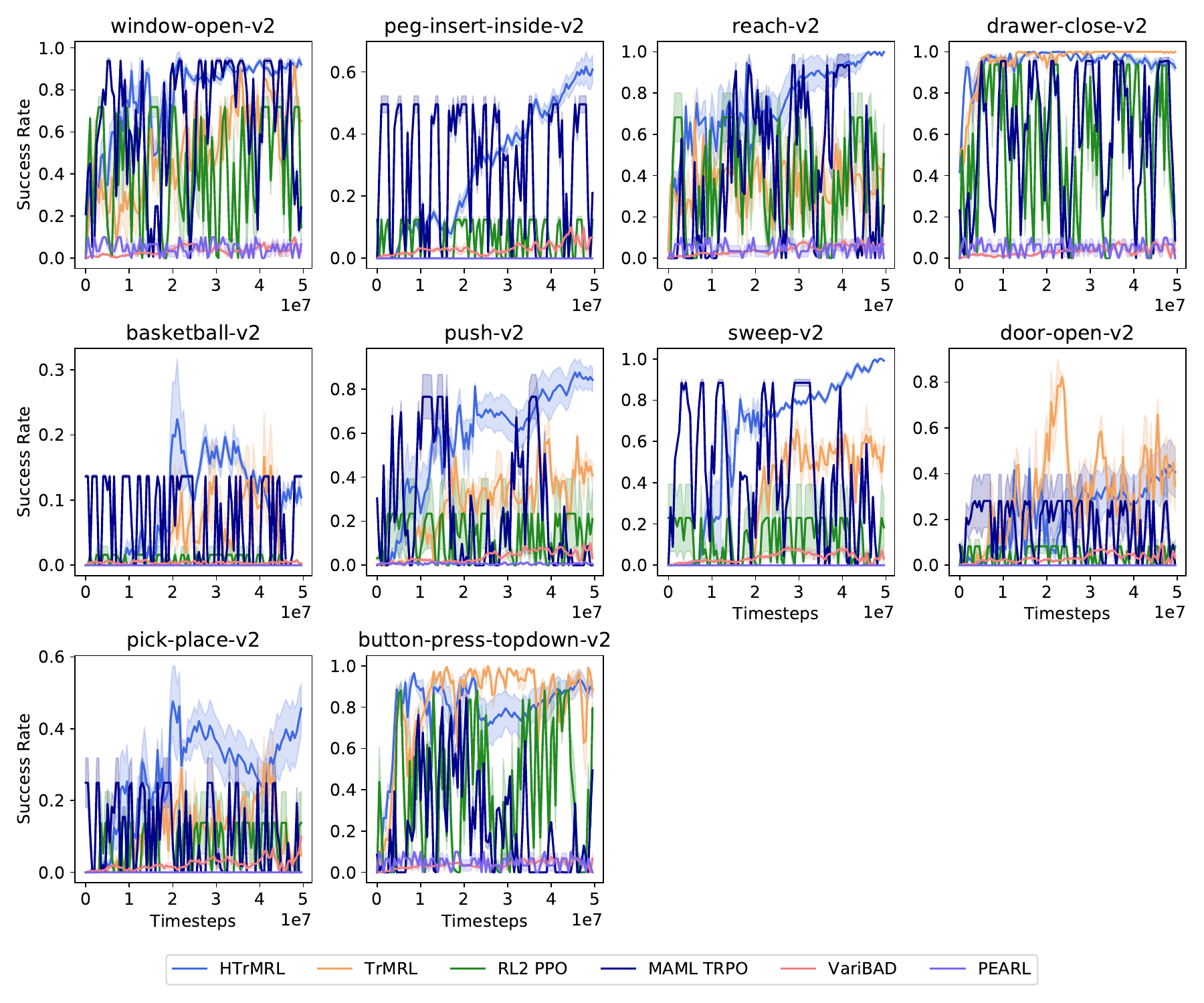}
    \caption{Success rate per train task for the ML10 benchmark for HTrMRL, TrMRL, PEARL, RL2 PPO, MAML TRPO, and VariBAD.}
    \label{fig:plot_ml10_average_success_rate_train_task}
\end{figure*}

\begin{figure*}[htbp!]
    \centering
    \includegraphics[width=\textwidth]{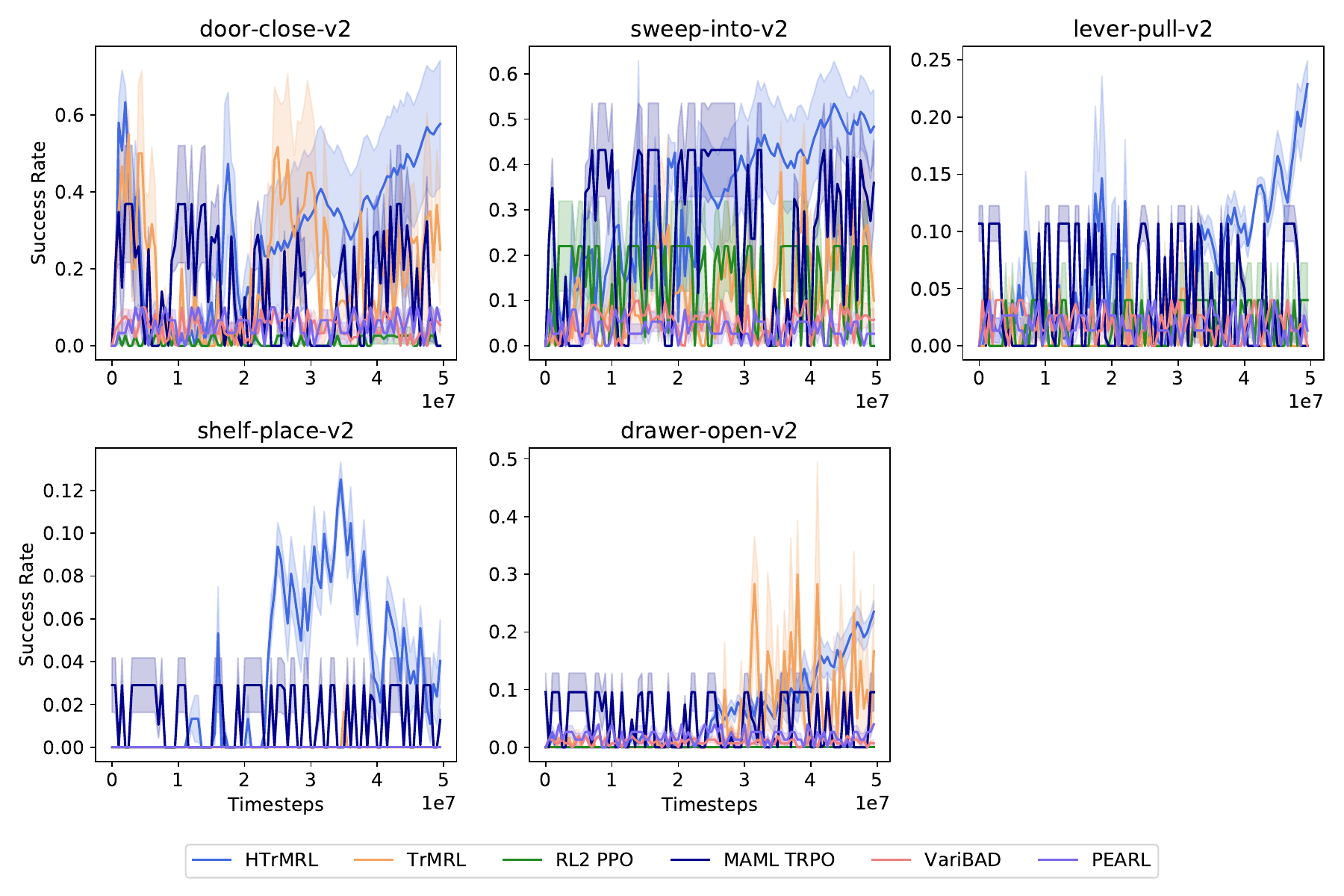}
    \caption{Success rate per test task for the ML10 benchmark for HTrMRL, TrMRL, PEARL, RL2 PPO, MAML TRPO, and VariBAD.}
    \label{fig:plot_ml10_average_success_rate_test_task}
\end{figure*}

\begin{figure*}[htbp]
    \centering
    \includegraphics[width=0.85\textwidth]{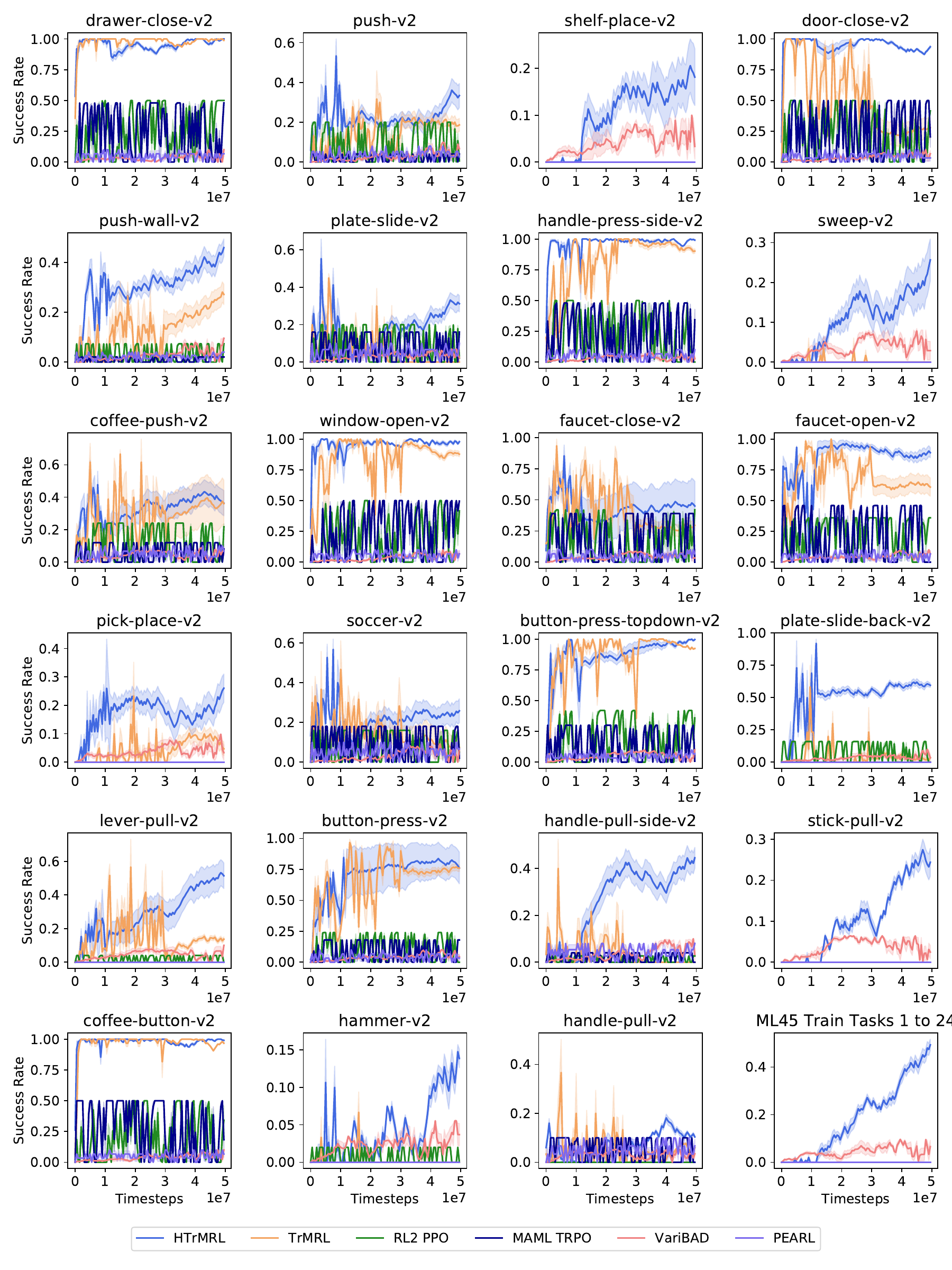}
    \caption{Success rate per train task for the ML45 benchmark for HTrMRL, TrMRL, PEARL, RL2 PPO, MAML TRPO, and VariBAD (Part 1).}
    \label{fig:plot_ml45_average_success_rate_train_task_1}
\end{figure*}

\begin{figure*}[htbp]
    \centering
    \includegraphics[width=0.85\textwidth]{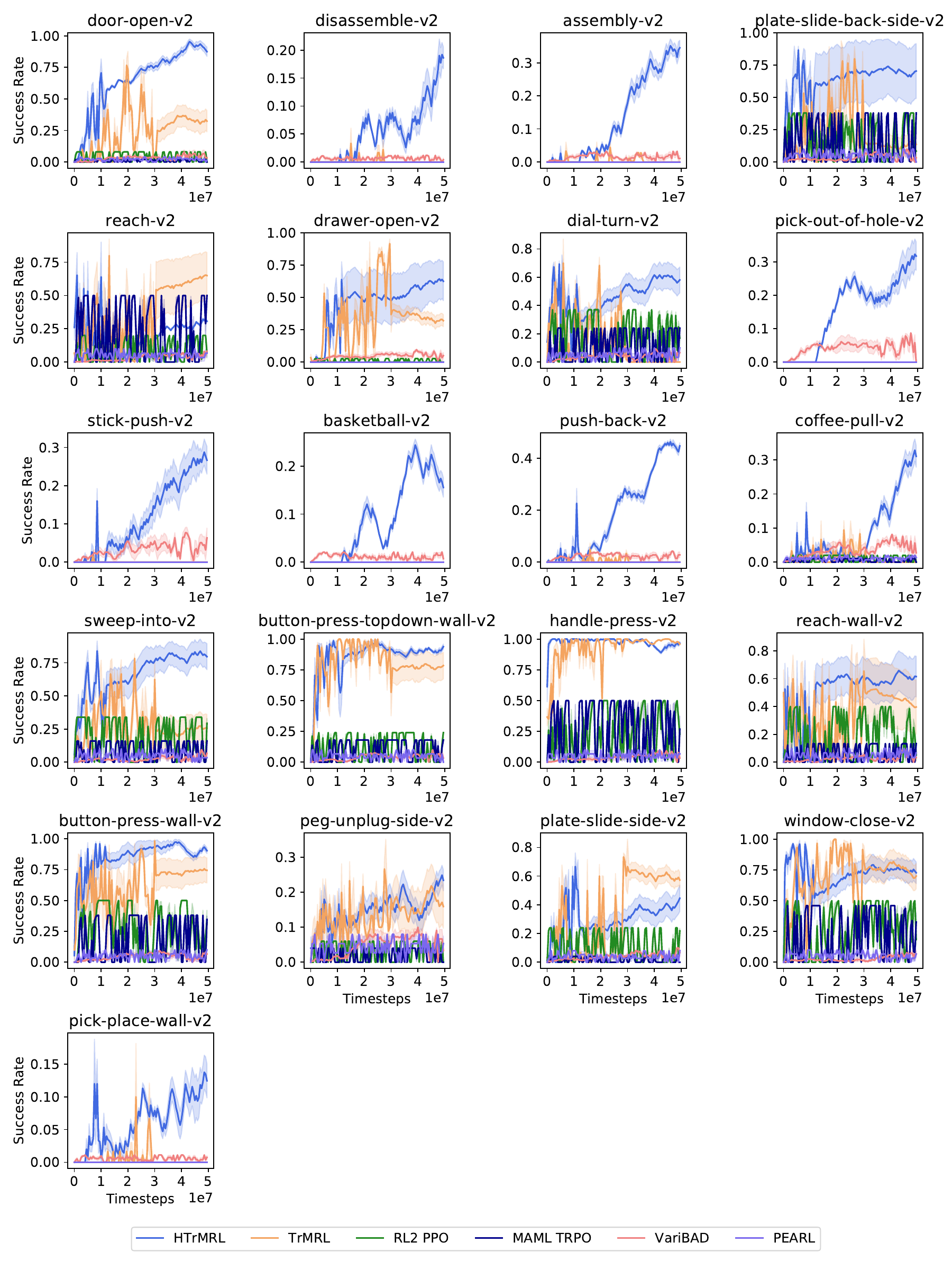}
    \caption{Success rate per train task for the ML45 benchmark for HTrMRL, TrMRL, PEARL, RL2 PPO, MAML TRPO, and VariBAD (Part 2).}
    \label{fig:plot_ml45_average_success_rate_train_task_2}
\end{figure*}

\begin{figure*}[htbp]
    \centering
    \includegraphics[width=\textwidth]{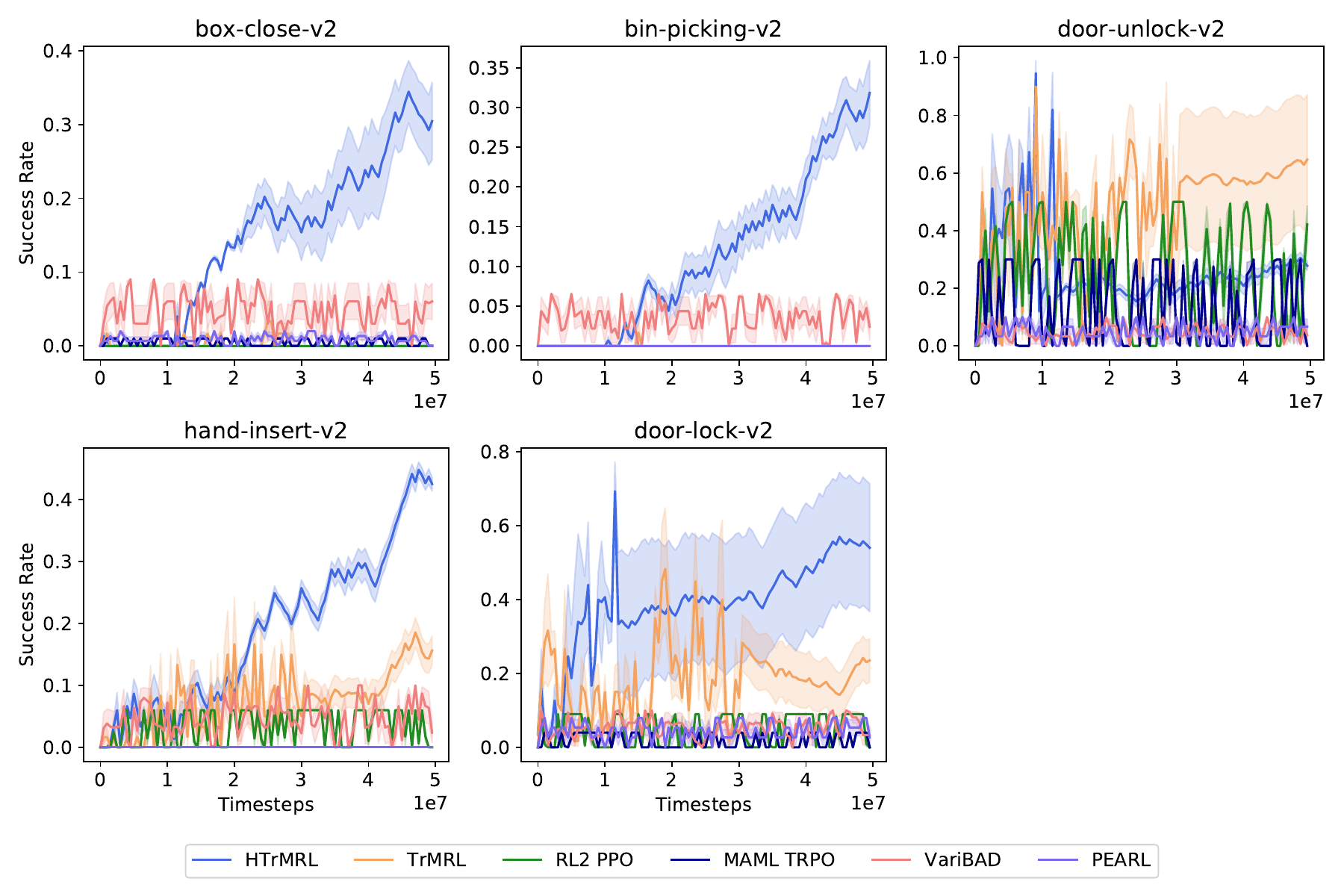}
    \caption{Success rate per test task for the ML45 benchmark for HTrMRL, TrMRL, PEARL, RL2 PPO, MAML TRPO, and VariBAD.}
    \label{fig:plot_ml45_average_success_rate_test_task}
\end{figure*}

Through investigating the individual plots for each task in terms of average success rate in Figures~\ref{fig:plot_ml10_average_success_rate_train_task}-~\ref{fig:plot_ml45_average_success_rate_test_task}, the difference in the level of difficulty for the tasks can be demonstrated, noticing that e.g. in Figure~\ref{fig:plot_ml10_average_success_rate_test_task} the \textit{Door Close} task is easier to successfully perform after meta-training, reaching success rates of approximately $0.6$, compared to the \textit{Shelf Place} task, for which all the methods fail to reach a success rate higher than $0.12$. This is also the case for the performance per task for the test tasks in ML45, which we show in Figure~\ref{fig:plot_ml45_average_success_rate_test_task}. For the \textit{Bin Picking} task, none of the baselines can successfully finish even  one episode, except for VariBAD with success rate of approximately $0.05$ (HTrMRL being more robust, reaching $0.3$). On the other hand, for the \textit{Door Unlock} task, RL2 PPO manages to reach success rates of approximately $0.4$, and TrMRL reaches $0.6$, being one of the few cases where HTrMRL underperforms by reaching a success rate of $0.3$.

\begin{figure*}[ht]%
    \centering%
%
    \begin{minipage}{0.49\textwidth}%
        \centering%
        \includegraphics[width=1.2\textwidth]{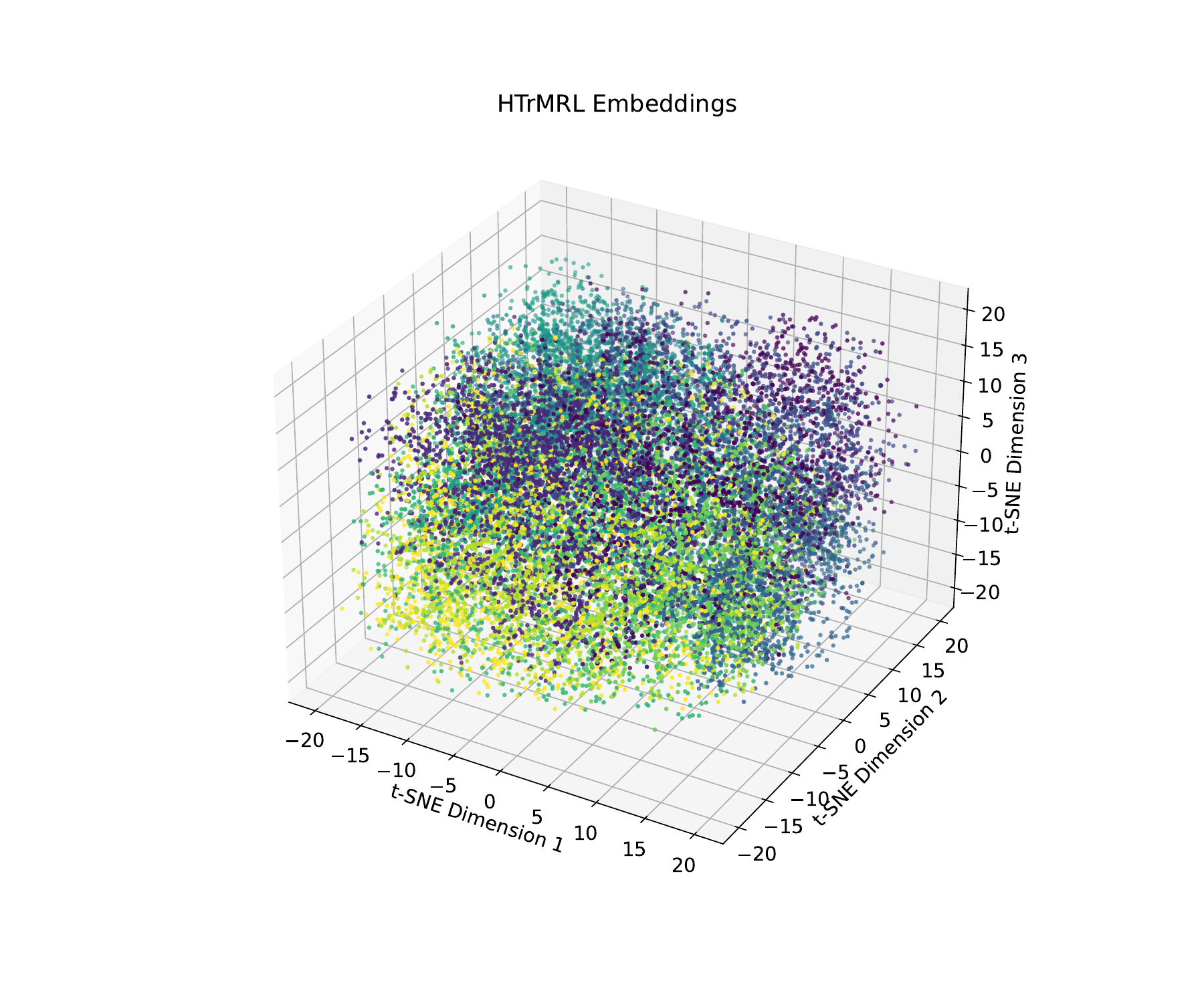}%
    \end{minipage}%
    \begin{minipage}{0.49\textwidth}%
        \centering%
        \hspace{-2em}%
        \includegraphics[width=1.2\textwidth]{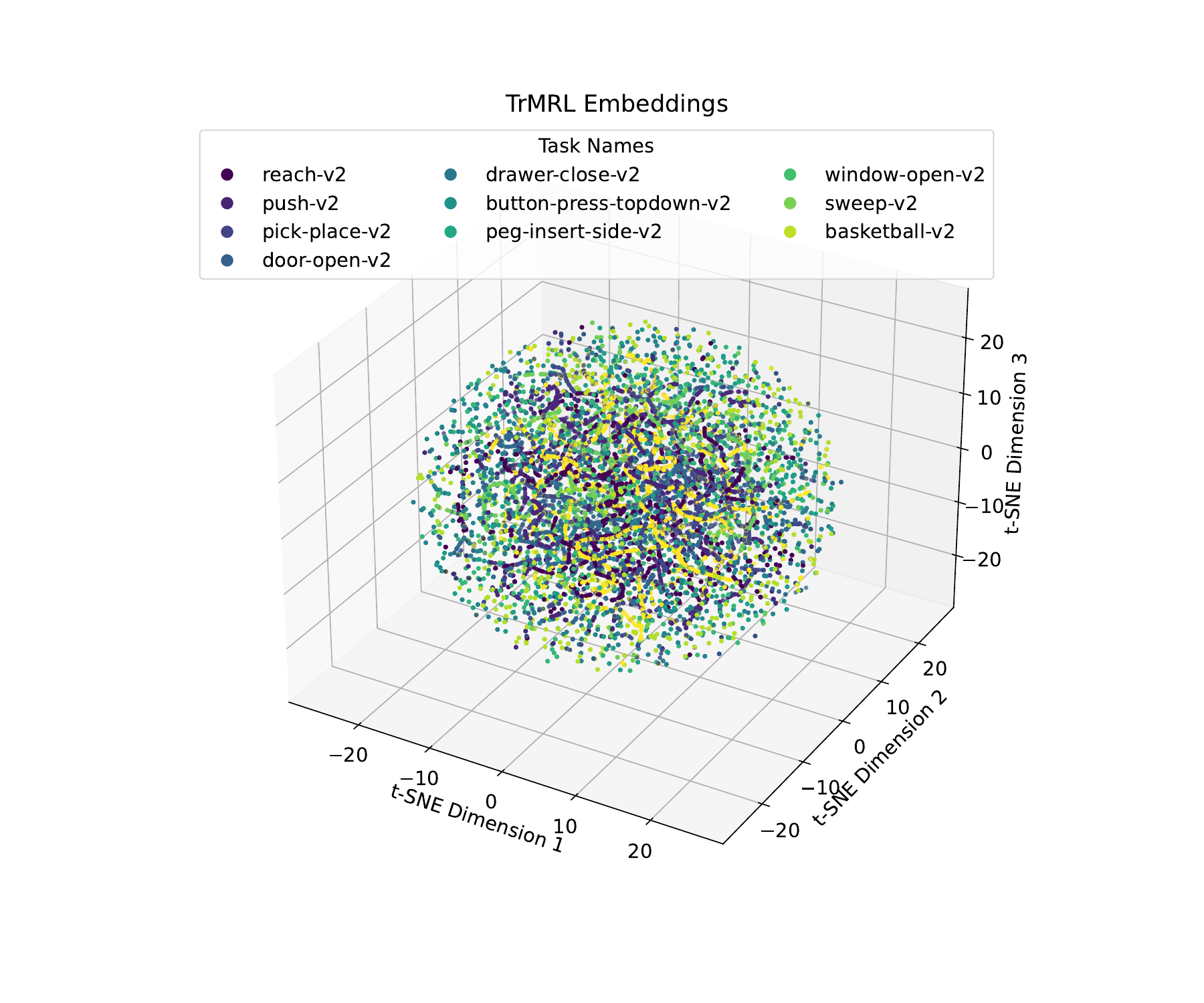}%
    \end{minipage}%
    \vspace{-2em}%
    \caption{T-SNE plots of the output embeddings for HTrMRL(left) and TrMRL(right) for the tasks in the training set of the ML10 benchmark of Meta-World.}%
    \label{fig:TSNE_plots_all}%
\end{figure*}
\begin{figure*}[htbp]
    \centering
    \includegraphics[width=\textwidth]{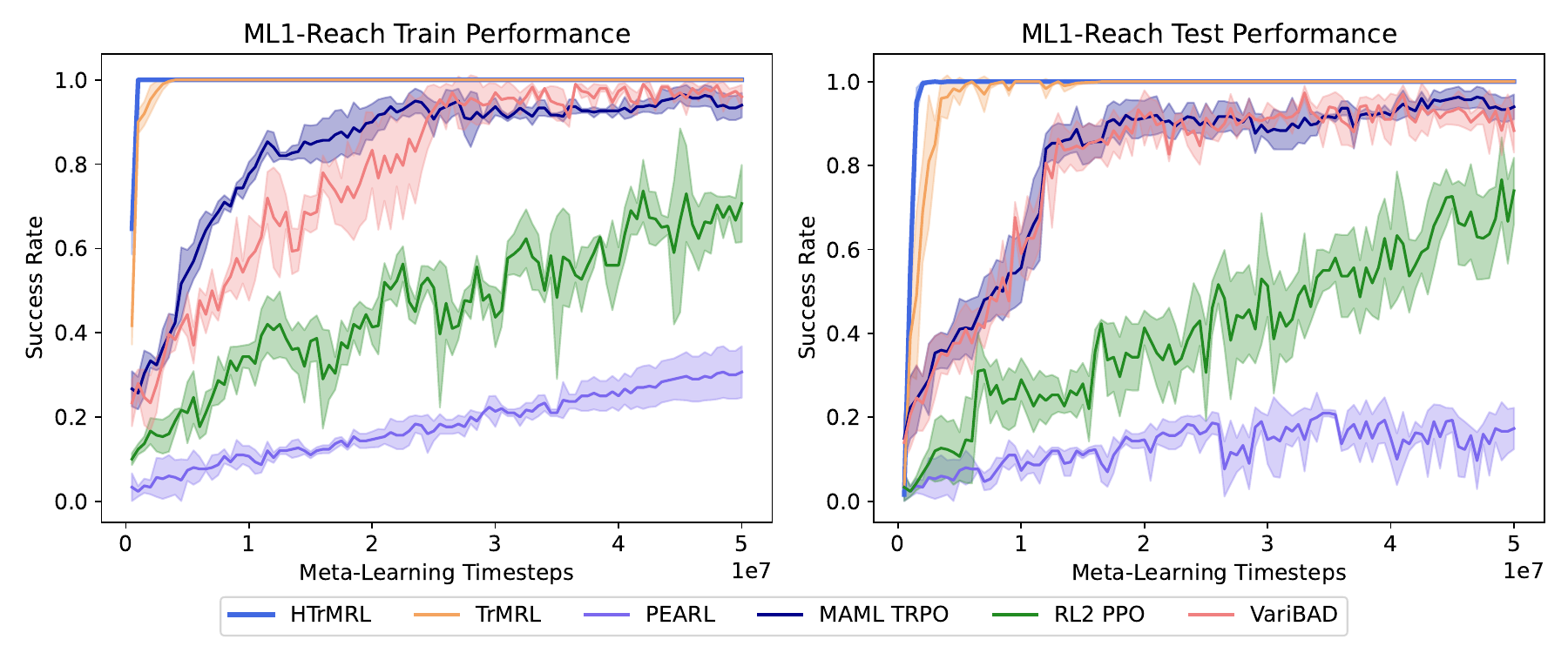}
    \caption{Meta-Train and Test performance in terms of Average Success Rate of HTrMRL, TrMRL, PEARL, MAML TRPO, RL2 PPO and VariBAD on the ML1 benchmark for training(left) and testing(right) on parametric variations of the \textit{Reach} task.}
    \label{fig:plot_ml1_reach_average_success}
\end{figure*}
\begin{figure*}[htbp]
    \centering
    \includegraphics[width=\textwidth]{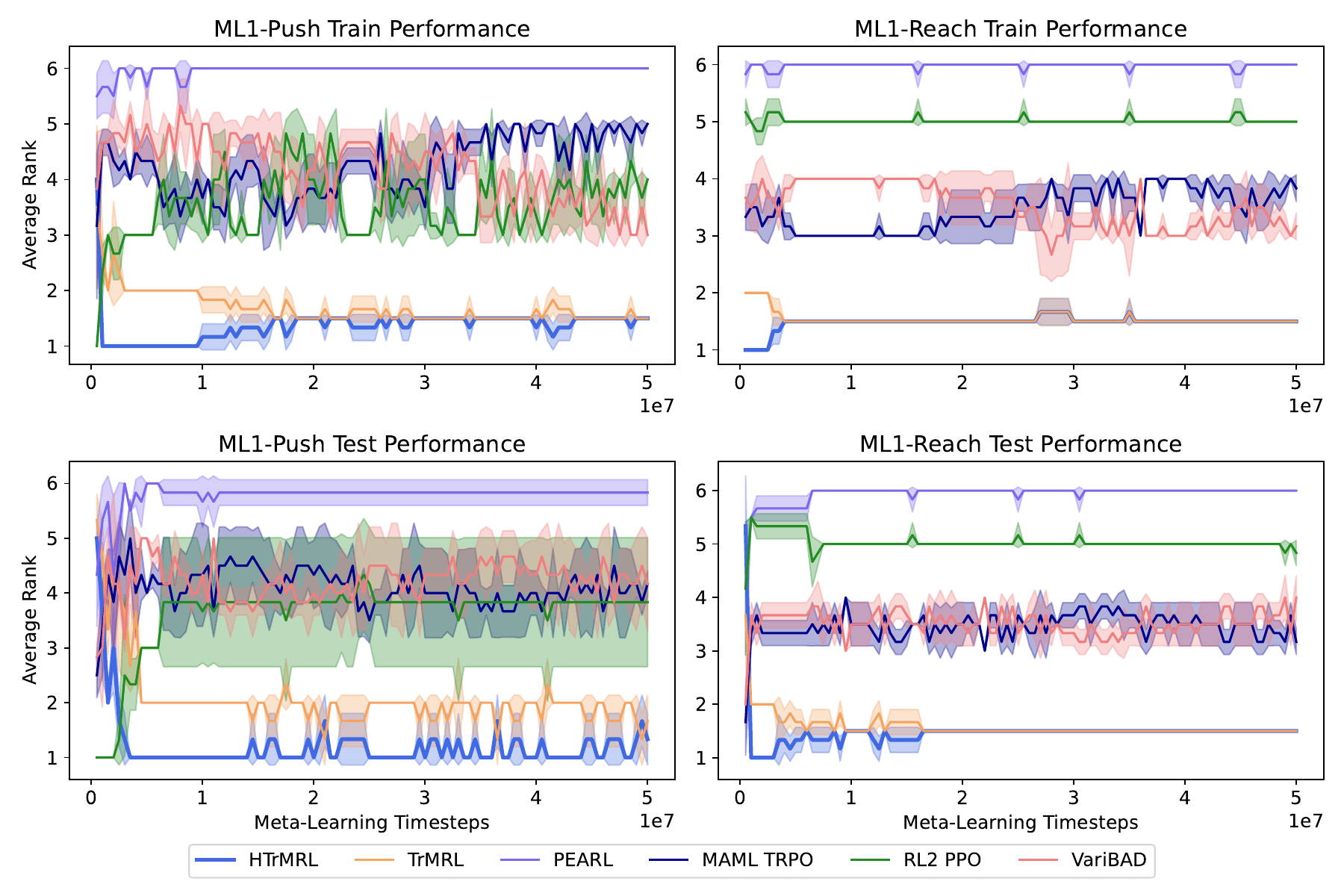}
    \caption{Average ranking plots of Meta-Train and Test performance for all methods on the ML1 \textit{Push}(left) and \textit{Reach}(right). Lower ranks indicate better performance.}
    \label{fig:plot_ml1_ranks}
\end{figure*}

\begin{figure*}[ht!]
    \centering
    \includegraphics[width=0.5\textwidth]{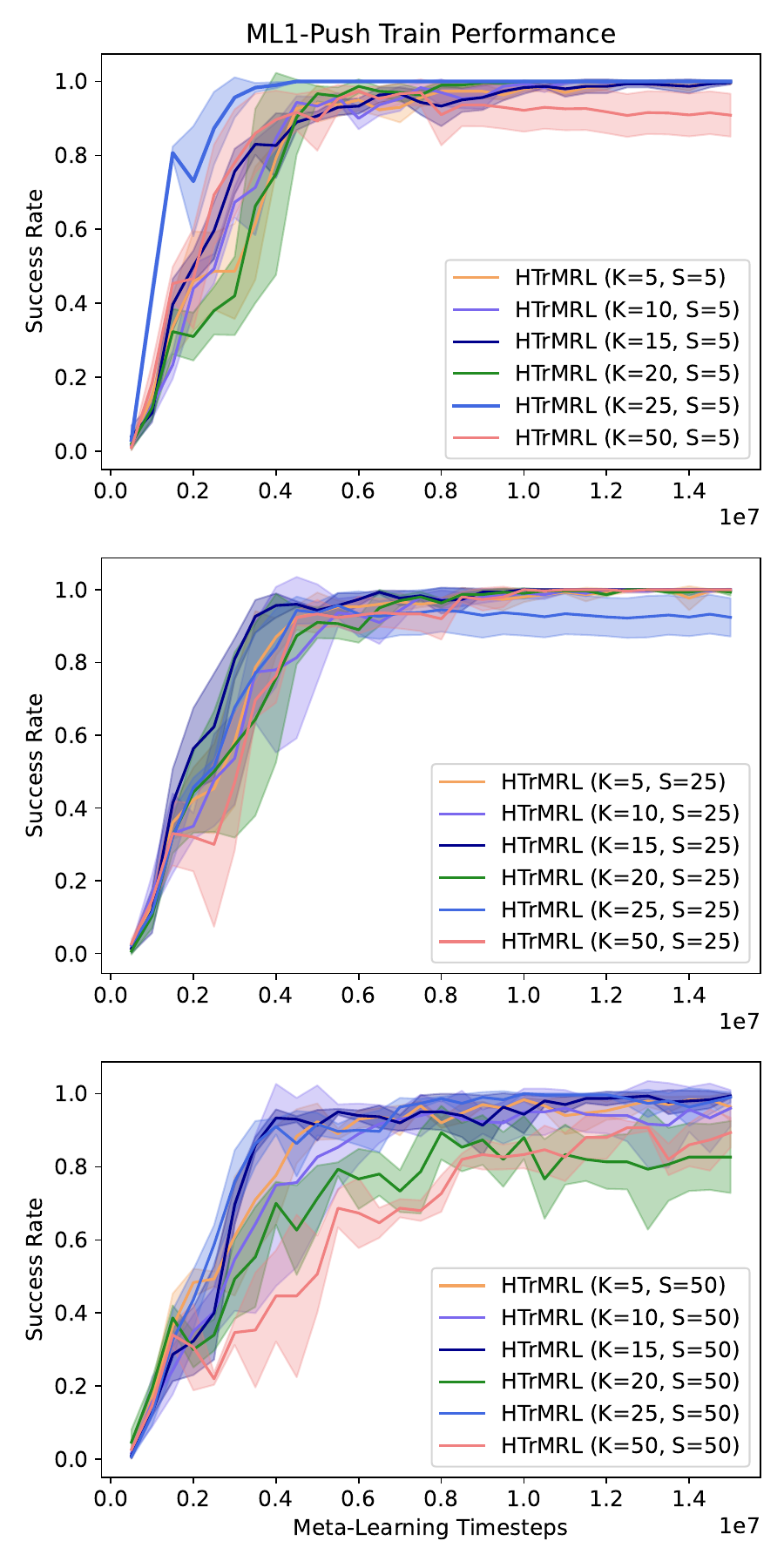}
    \caption{Investigating the meta-training performance impact of the sequence length S and number of episodes K for HTrMRL in ML1 \textit{Push}. The top plot shows the performance of sequence lengths $S=5$ with different episode lengths $K \in \{5, 10, 15, 20, 25, 50\}$. The middle plot shows the performance of $S=25$ with $K \in \{5, 10, 15, 20, 25, 50\}$. For the bottom plot, $S=50$.}
    \label{fig:plot_ml1_e_and_s_ablation}
\end{figure*}

\begin{figure*}[ht!]
    \centering
    \includegraphics[width=0.5\textwidth]{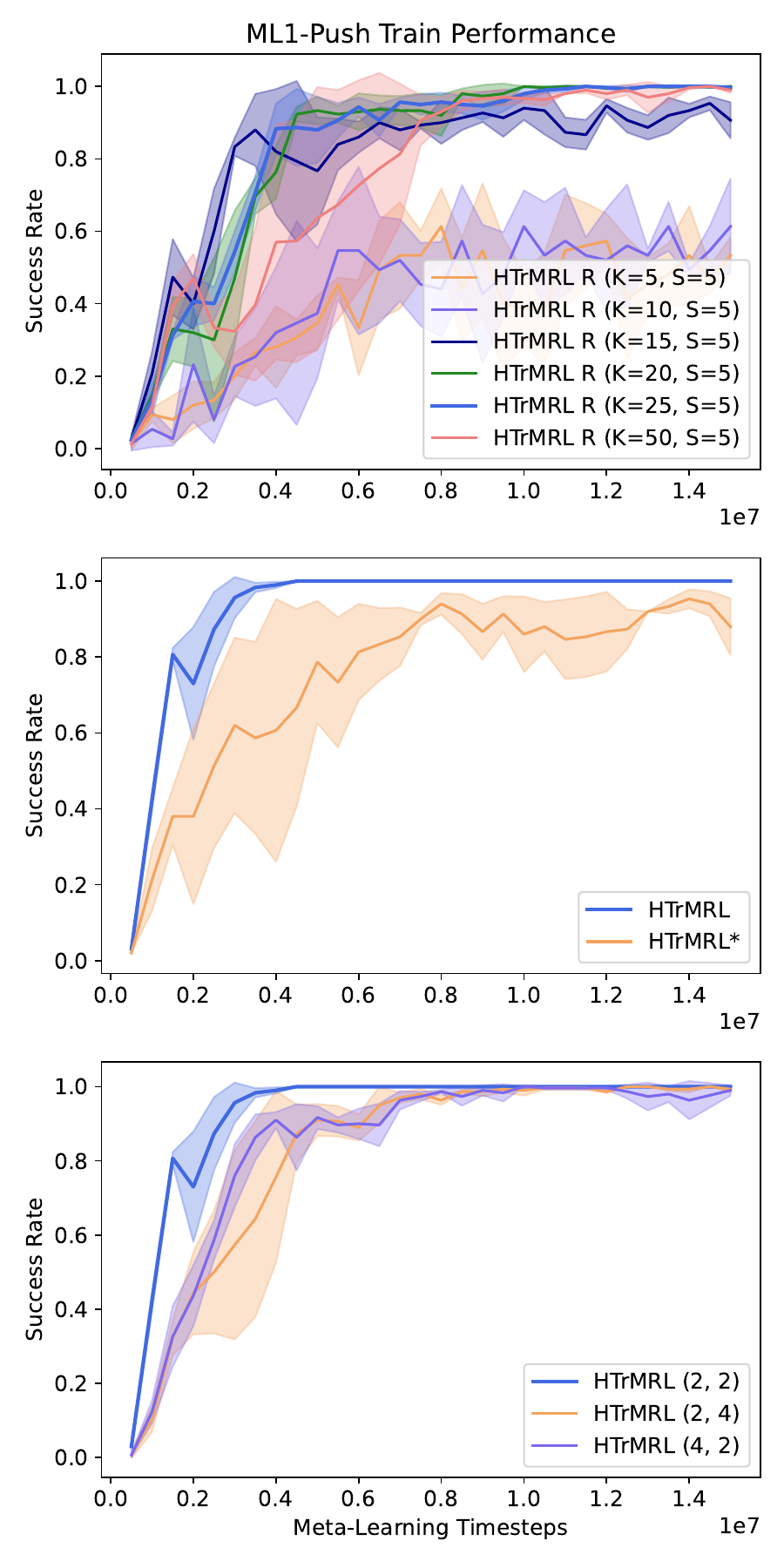}
    \caption{Investigating the meta-training performance impact of the sampling strategy for the sequence for HTrMRL in ML1 \textit{Push}. The top plot shows the performance of HTrMRL when sampling $K$ sequences from random episodes (i.e. more than one sequence per episode). The middle plot shows the performance impact of concatenating a linear transformation of the state(HTrMRL) vs using only the output of the transformer(HTrMRL*) as the input for the policy. The bottom plot shows the performance impact of the number of encoder blocks, comparing configurations of $2$ encoder blocks for each $\mathbf{T}_{1}$ and $\mathbf{T}_{2}$, $2$ encoder blocks for $\mathbf{T}_{1}$ and $4$ for $\mathbf{T}_{2}$,
    and $4$ encoder blocks for $\mathbf{T}_{1}$ and $2$ for $\mathbf{T}_{2}$.}
    \label{fig:plot_ml1_r_s_and_a_ablation}
\end{figure*}

\begin{figure*}[htbp]
    \centering
    \includegraphics[width=\textwidth]{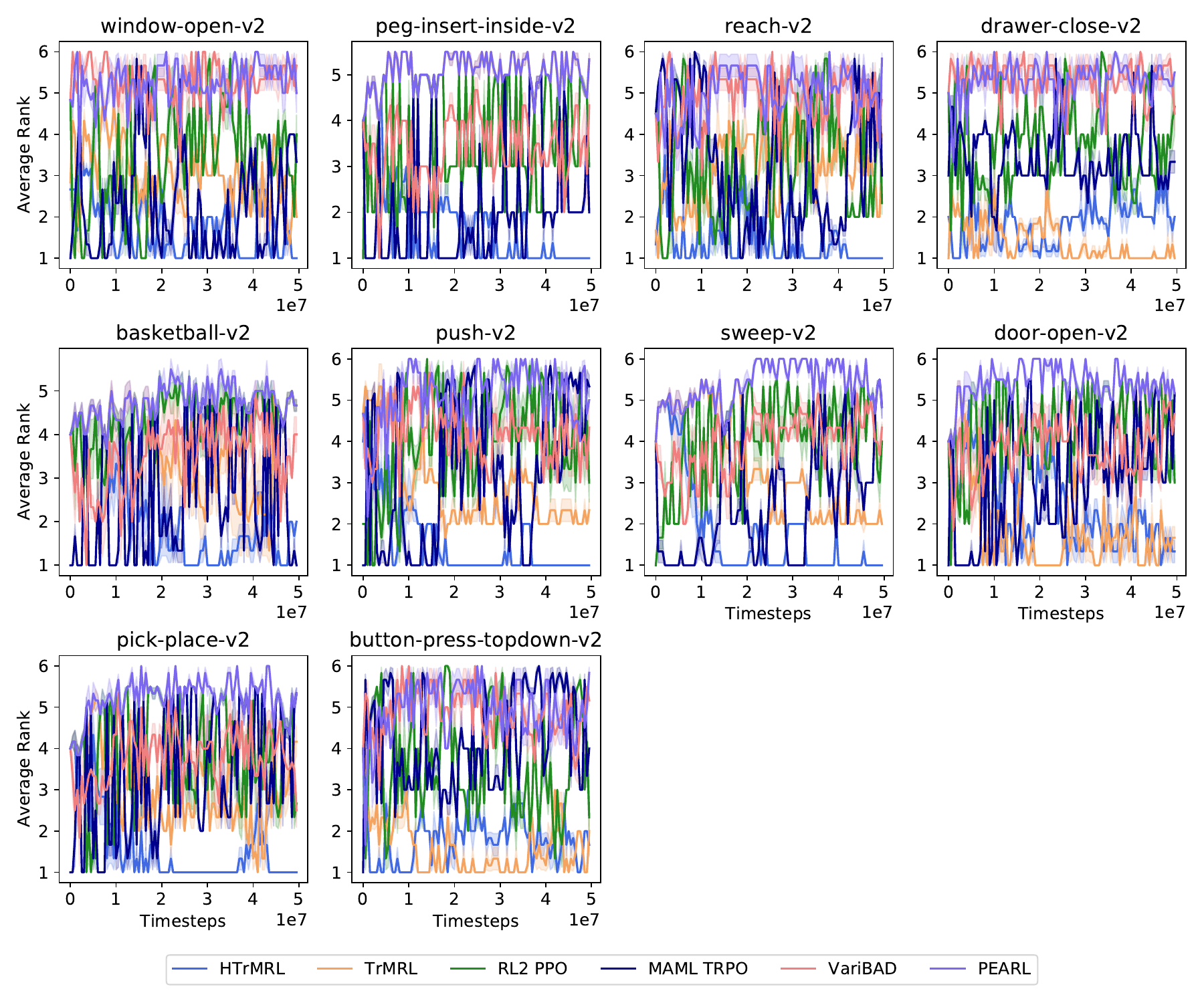}
    \caption{Average rank per train task for the ML10 benchmark for HTrMRL, TrMRL, PEARL, RL2 PPO, MAML TRPO, and VariBAD.}
    \label{fig:plot_ml10_rank_train_task}
\end{figure*}
\begin{figure*}[htbp]
    \centering
    \includegraphics[width=\textwidth]{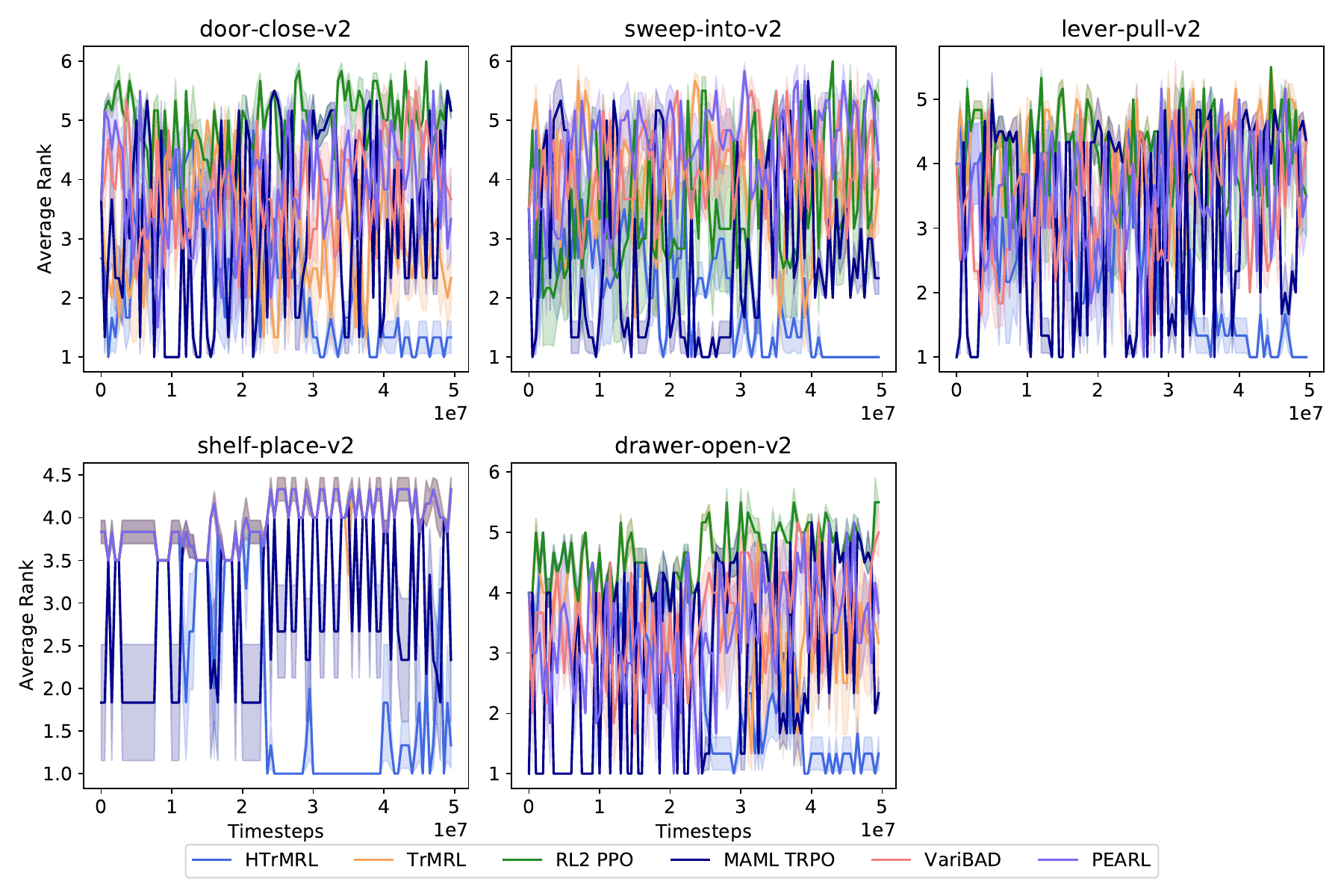}
    \caption{Average rank per test task for the ML10 benchmark for HTrMRL, TrMRL, PEARL, RL2 PPO, MAML TRPO, and VariBAD.}
    \label{fig:plot_ml10_rank_test_task}
\end{figure*}
\begin{figure*}[htbp]
    \centering
    \includegraphics[width=0.85\textwidth]{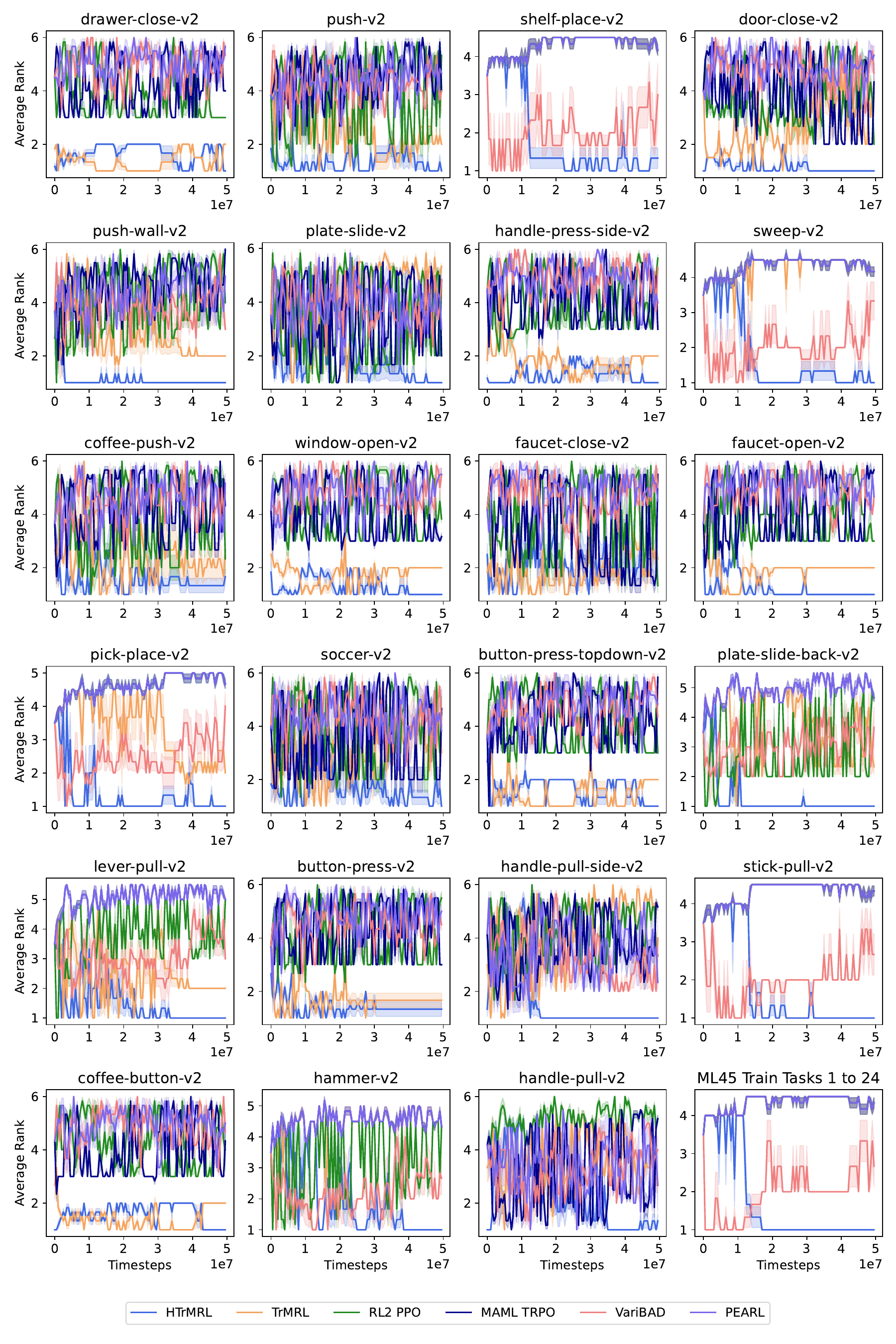}
    \caption{Average rank per train task for the ML45 benchmark for HTrMRL, TrMRL, PEARL, RL2 PPO, MAML TRPO, and VariBAD (Part 1).}
    \label{fig:plot_ml45_rank_train_task_1}
\end{figure*}
\begin{figure*}[htbp]
    \centering
    \includegraphics[width=0.85\textwidth]{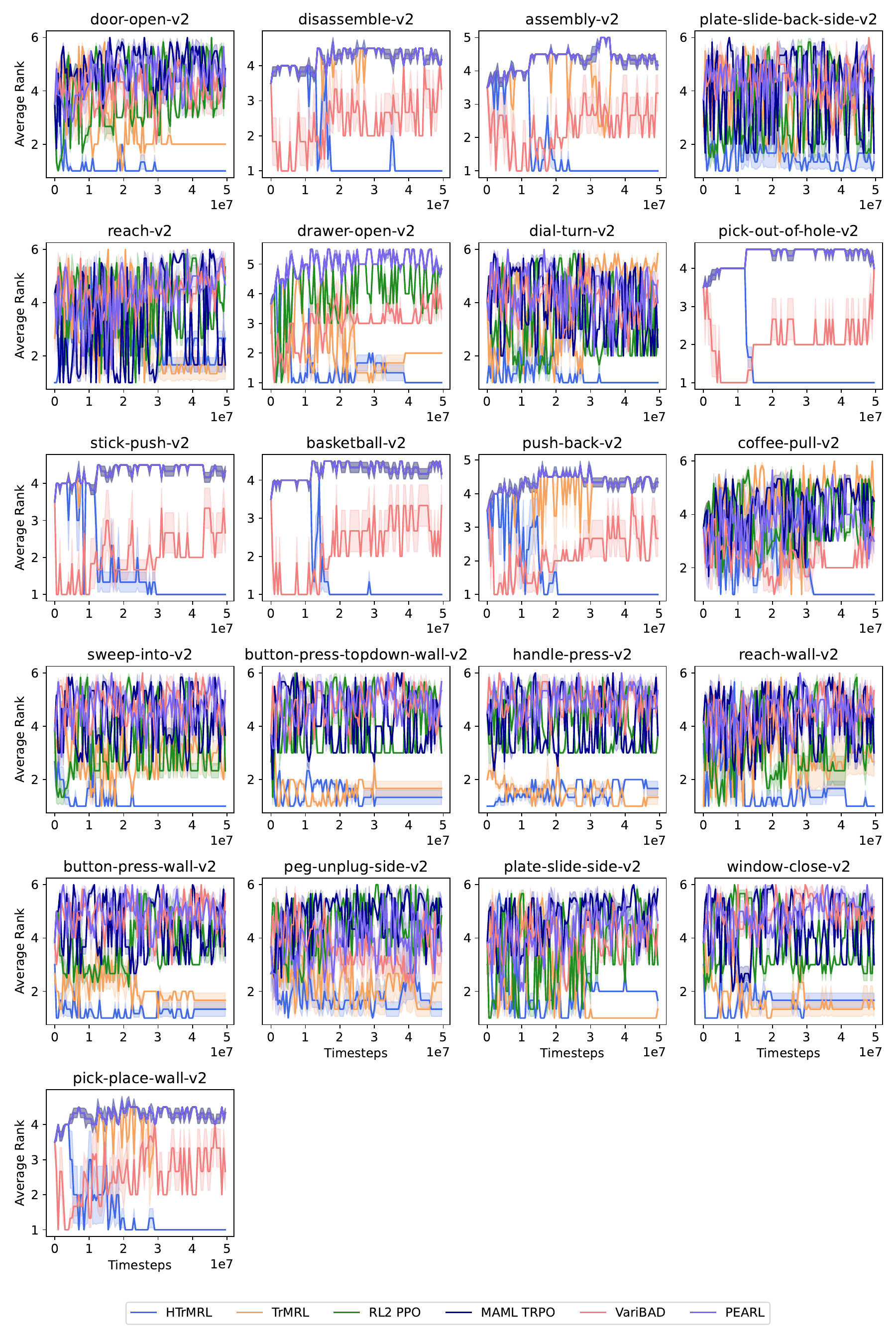}
    \caption{Average rank per train task for the ML45 benchmark for HTrMRL, TrMRL, PEARL, RL2 PPO, MAML TRPO, and VariBAD (Part 2).}
    \label{fig:plot_ml45_rank_train_task_2}
\end{figure*}
\begin{figure*}[htbp]
    \centering
    \includegraphics[width=\textwidth]{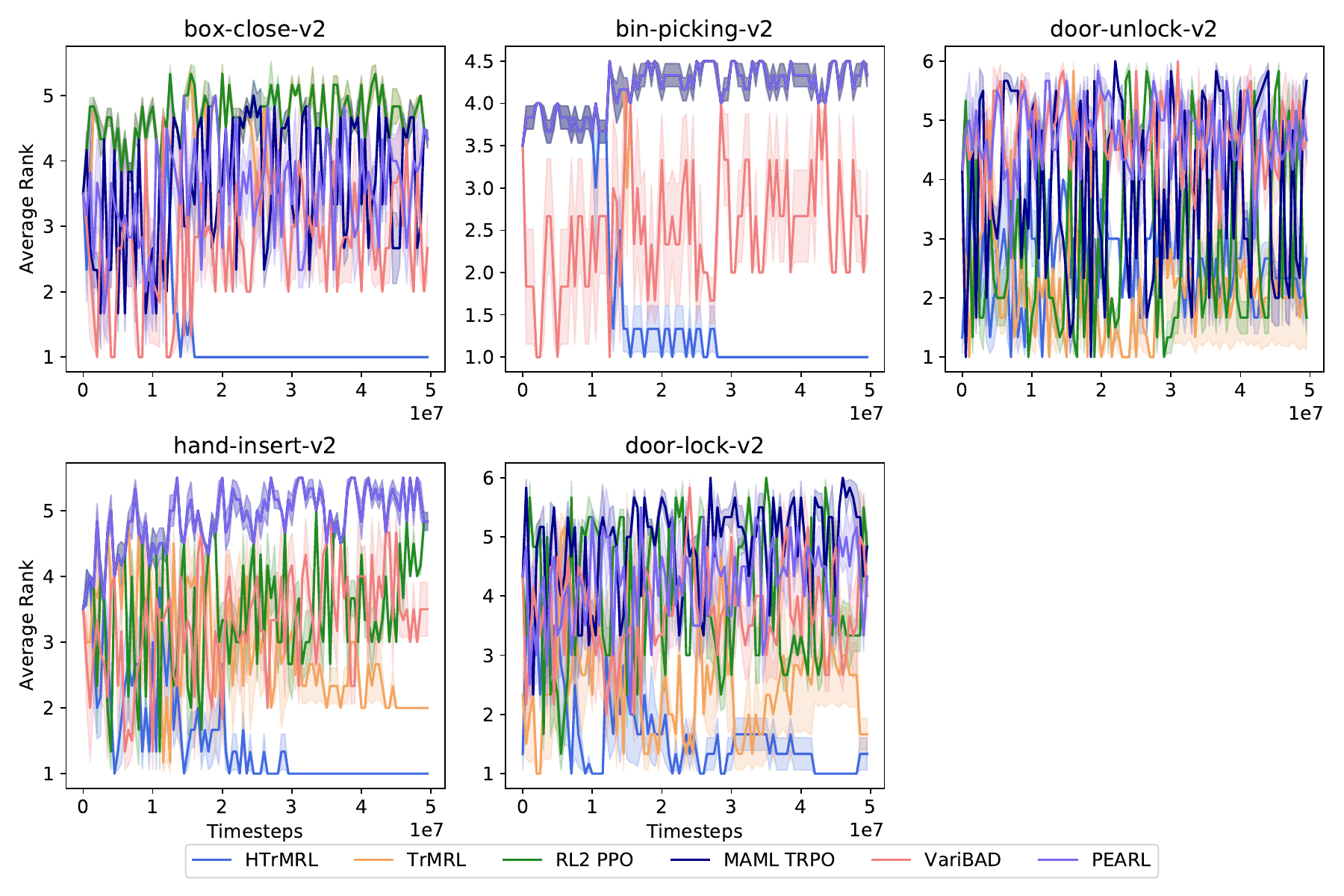}
    \caption{Average rank per test task for the ML45 benchmark for HTrMRL, TrMRL, PEARL, RL2 PPO, MAML TRPO, and VariBAD.}
    \label{fig:plot_ml45_rank_test_task}
\end{figure*}